\documentclass[lettersize,journal]{IEEEtran}
\usepackage{amsmath,amsfonts}
\usepackage{algorithmic}
\usepackage{algorithm}
\usepackage{array}
\usepackage[caption=false,font=normalsize,labelfont=sf,textfont=sf]{subfig}
\usepackage{textcomp}
\usepackage{stfloats}
\usepackage{url}
\usepackage{verbatim}
\usepackage{hyperref}
\usepackage{graphicx}
\usepackage{cite}
\usepackage{multirow}
\usepackage{float}
\usepackage{array}
\usepackage{pifont}
\usepackage{booktabs}
\usepackage{xcolor}
\usepackage{makecell}
\usepackage{pifont}
\usepackage[table]{xcolor}
\definecolor{suppblue}{RGB}{53, 124, 188}
\definecolor{lightgray}{gray}{0.93}
\newcommand{\Checkmark}{\ding{51}}
\newcommand{\Xmark}{\ding{55}}
\newcolumntype{C}[1]{>{\centering\arraybackslash}p{#1}}

\begin{document}

\title{FloorPlan-VLN: A New Paradigm for Floor Plan Guided Vision-Language Navigation}

\author{Kehan Chen,
Yan Huang$^\dag$,
Dong An,
Jiawei He,
Yifei Su,
Jing Liu,
Nianfeng Liu,
Liang Wang\thanks{$^\dag$ Liang Wang and Yan Huang are the corresponding authors.}$^\dag$
\IEEEcompsocitemizethanks{
\IEEEcompsocthanksitem 
Kehan Chen, Yan Huang, and Liang Wang are with New Laboratory of Pattern Recognition (NLPR), Institute of Automation, Chinese Academy of Sciences and School of Artificial Intelligence, University of Chinese Academy of Sciences, China. Email: 
kehan.chen@cripac.ia.ac.cn, yhuang@nlpr.ia.ac.cn, wangliang@nlpr.ia.ac.cn.
Dong An is with AMap, Alibaba Group.
Email: andong979797@gmail.com.
Jiawei He is with BAAI.
Email: jwhe@baai.ac.cn.
Yifei Su is with Xiaomi Robotics Lab.
Email: suyifei@xiaomi.com.
Jing Liu and Nianfeng Liu are with FiveAges. Email: liujing0@mail.ustc.edu.cn and liunianfeng@five-ages.com.
}}

\markboth{Journal of \LaTeX\ Class Files,~Vol.~14, No.~8, August~2021}%
{Shell \MakeLowercase{\textit{et al.}}: A Sample Article Using IEEEtran.cls for IEEE Journals}


\maketitle

\begin{figure*}[!t]
    \centering
    \vspace{-1em}
    \includegraphics[width=\textwidth, trim=0 410 0 0, clip]{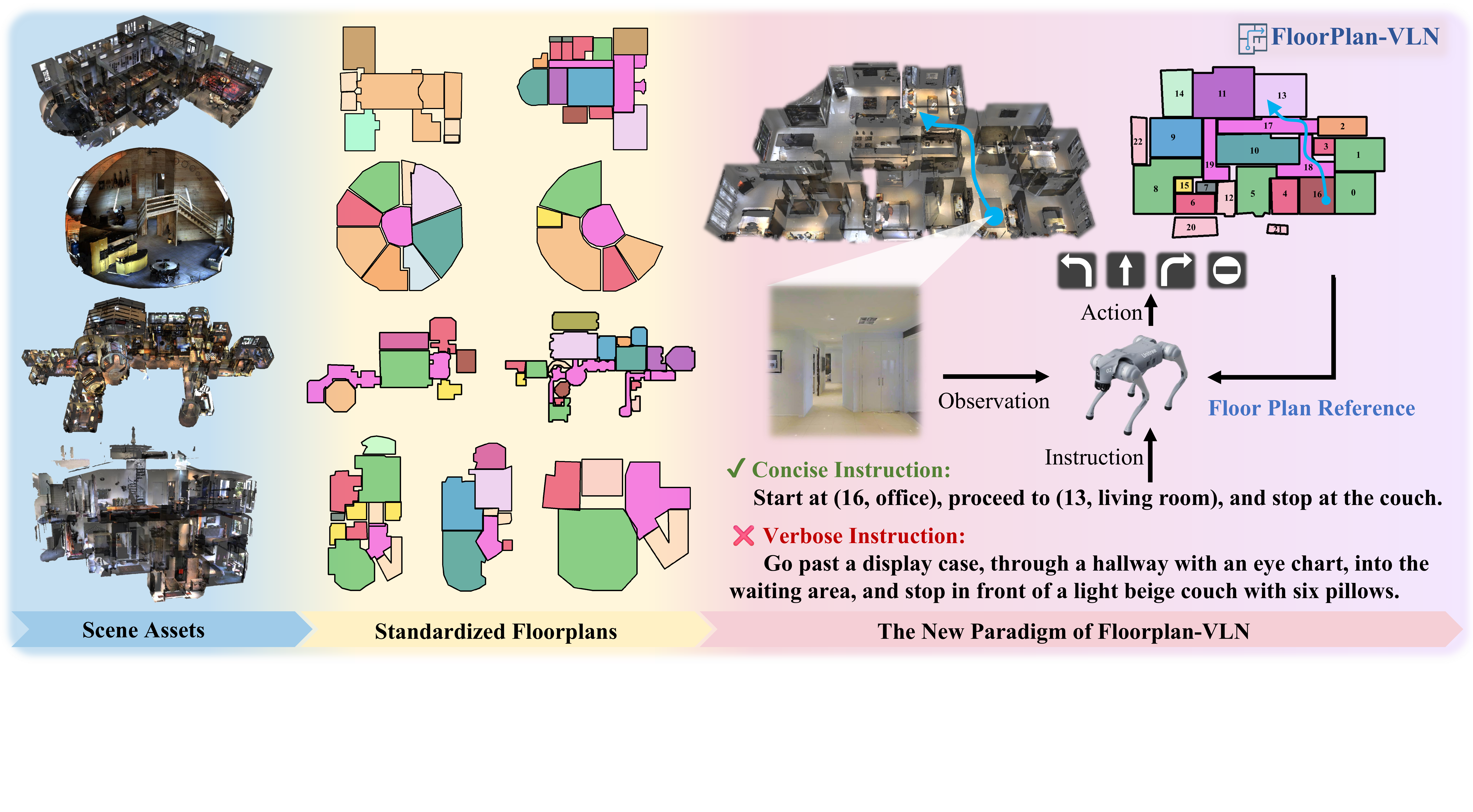}
    \vspace{-2.0em}
    \caption{Illustration of the FloorPlan-VLN paradigm. Floor plans are derived from Matterport3D scenes, and concise region-based instructions are constructed to train agents to navigate by leveraging floor plans as global spatial priors.}
    \label{fig:teaser}
    \vspace{-1em}
\end{figure*}

\begin{abstract}
Existing Vision-Language Navigation (VLN) task requires agents to follow verbose instructions, ignoring some potentially useful global spatial priors, limiting their capability to reason about spatial structures. 
Although human-readable spatial schematics (e.g., floor plans) are ubiquitous in real-world buildings, current agents lack the cognitive ability to comprehend and utilize them.
To bridge this gap, we introduce \textbf{FloorPlan-VLN}, a new paradigm that leverages structured semantic floor plans as global spatial priors to enable navigation with only concise instructions.
We first construct the FloorPlan-VLN dataset, which comprises over 10k episodes across 72 scenes. 
It pairs more than 100 semantically annotated floor plans with Matterport3D-based navigation trajectories and concise instructions that omit step-by-step guidance.
Then, we propose a simple yet effective method \textbf{FP-Nav} that uses a dual-view, spatio-temporally aligned video sequence, and auxiliary reasoning tasks to align observations, floor plans, and instructions.
When evaluated under this new benchmark, our method significantly outperforms adapted state-of-the-art VLN baselines, achieving more than a 60\% relative improvement in navigation success rate.
Furthermore, comprehensive noise modeling and real-world deployments demonstrate the feasibility and robustness of FP-Nav to actuation drift and floor plan distortions.
These results validate the effectiveness of floor plan guided navigation and highlight FloorPlan-VLN as a promising step toward more spatially intelligent navigation\footnote{Project page:~\href{https://github.com/Chenkehan21/FloorPlan-VLN/tree/main}{FloorPlan-VLN}}.
\end{abstract}


\section{Introduction}
\label{sec:intro}
\IEEEPARstart{V}{ision}-Language Navigation (VLN)~\cite{anderson2018vision} has long been a core challenge in embodied AI, requiring agents to navigate unseen environments following linguistic instructions.
Despite significant progress, the existing VLN paradigm heavily relies on verbose, step-by-step instructions, which imposes a high cognitive burden on human-robot interaction.
This reliance stems from a critical limitation: the absence of global spatial priors. 
In our daily lives, human-readable global schematics, such as floor plans in shopping malls and hospitals, are widely available. 
Humans exhibit remarkable spatial intelligence by intuitively interpreting such diagrams, enabling them to rapidly plan routes and produce concise instructions that specify only the start region, target region and stopping conditions. 
In contrast, existing VLN agents pay little attention to these readily available resources. 

Inspired by this, we argue that robots should learn to exploit floor plans when available. 
Unlike dense metric maps that require extensive SLAM~\cite{xu2021fastlio} or 3D reconstruction~\cite{mildenhall2021nerf}, floor plans are lightweight, highly accessible and remain invariant to object placements.
Motivated by this, we introduce \textbf{FloorPlan-VLN}, a novel paradigm that offloads spatial complexity from textual instructions onto global floor plan priors, as shown in Figure~\ref{fig:teaser}. 
This shift enables a more natural human-robot interaction via concise instructions which solely refer to the start and target regions of the floor plans, as well as the stop conditions derived from human intentions.
In this setting, agents must bridge the semantic gap between an abstract floor plan and egocentric observations to execute concise instructions.
Crucially, we define the floor plan as a \textbf{Unified Spatial Interface} that can be derived from various sources, ranging from precise architectural blueprints to informal, hand-drawn sketches provided by users. 
This flexibility ensures that our paradigm is practical in diverse real-world scenarios where high-fidelity maps may be unavailable.

However, this new paradigm has rarely been investigated and first faces the challenge of data scarcity.
Existing VLN datasets~\cite{anderson2018vision, ku2020rxr, krantz_vlnce_2020} only pair navigation instructions with visual trajectories, lacking floor plan metadata.
To bridge this gap, we construct the FloorPlan-VLN dataset, which pairs semantically annotated floor plans with navigation trajectories and concise instructions.
The floor plans are derived from Matterport3D~\cite{chang2017matterport3d} and enriched with region types and identifiers.
A large number of trajectories are further annotated at the waypoint level, where each waypoint is assigned a region identifier and semantic type.
This allows us to trace the agent’s movement across regions throughout the trajectory, which is crucial for filtering valid paths, generating concise instructions, and supporting localization-related auxiliary tasks.
To our knowledge, this is the first benchmark for studying navigation under floor plan priors with concise instructions.

The subsequent challenge is the alignment between abstract floor plans, egocentric observations and instructions. 
As floor plans provide only coarse global layouts rather than navigable metric maps, agents must learn to align global spatial priors and local observations.
Recently, Multimodal Large Language Models (MLLMs)~\cite{wang2024qwen2, bai2025qwen2} have demonstrated exceptional capabilities in interpreting 2D diagrams and schematics due to web-scale pretraining. 
To leverage its alignment ability for navigation tasks, we propose \textbf{FP-Nav}, a MLLM-based framework.
Specifically, we rasterize floor plans into images augmented with visual prompt marks, and introduce a dual-view, spatio-temporally aligned video pairing strategy.
This strategy synchronizes observations (view 1) with dynamic floor plan visualizations (view 2), ensuring a consistent correspondence between local observations and global layouts.
In addition, we jointly train three auxiliary tasks, namely region localization, trajectory reasoning, and instruction summarization, to enhance the model’s ability to align fine-grained observations, coarse-grained floor plans, and instructions.

When evaluated under this new benchmark, FP-Nav significantly outperforms adapted state-of-the-art (SoTA) VLN baselines.
Compared with the baseline finetuned on the FloorPlan-VLN dataset, our method achieves 138\% and 69\% relative improvements in success rate on seen and unseen splits, respectively, highlighting the potential of leveraging spatial priors.
More importantly, we rigorously evaluate our framework beyond ideal simulation environments. 
Through comprehensive noise modeling and zero-shot deployment on a real-world quadruped robot, we demonstrate that FP-Nav is robust to moderate actuation drift and uncalibrated floor plan distortions. 
These results not only validate the effectiveness of our approach but also establish FloorPlan-VLN as a promising step toward spatially intelligent embodied navigation.

In summary, our main contributions are three-fold:
\begin{itemize}
\item \textbf{New paradigm:} We introduce a novel floor plan guided VLN paradigm where agents follow concise, goal-oriented instructions under global spatial priors.
\item \textbf{New Dataset:} We construct the FloorPlan-VLN dataset, pairing semantically annotated floor plans with trajectories whose waypoints are labeled with region identifiers and types, along with concise region-level instructions.
\item \textbf{Model and baseline:} We propose FP-Nav, a model based on MLLMs with a spatio-temporally aligned input strategy and auxiliary tasks for cross-modal alignment.
Extensive evaluations, including noise modeling, ablation studies and real-world deployment, establish it as the first strong and robust baseline for this new paradigm.
\end{itemize}

\section{Related Work}
\label{sec:related_work}
\begin{table*}[t]
\centering
\caption{Conceptual comparison between FloorPlan-VLN and related navigation paradigms. Our task introduces a unique combination of \textbf{concise language} and \textbf{semantic-rich floor plans}, shifting the focus to spatio-semantic reasoning.}
\vspace{-1em}
\label{tab:paradigm_comparison}
\resizebox{\textwidth}{!}{
\begin{tabular}{lccccc}
\toprule
\textbf{Task / Paradigm} & \textbf{Instruction Type} & \textbf{Spatial Prior Type} & \textbf{Semantic Info.} & \textbf{Goal Specification} & \textbf{Primary Challenge} \\ \midrule
Traditional VLN \cite{anderson2018vision} & Verbose (step-by-step) & \Xmark (Unknown) & \Xmark & Linguistic Landmark & Trajectory Grounding \\
Hand-Drawn Navigation \cite{tan2025mobile, xu2024robot} & \Xmark (None) & Informal Sketch & \Xmark & Trajectory Annotation & Informal Cue Interpretation \\
FloorPlan Point Navigation \cite{li2025flona, shen2025vf} & \Xmark (None) & Geometric Boundary & \Xmark & Point Goal Coordinates & Localization \& Planning \\ \midrule
\rowcolor{lightgray} 
\textbf{FloorPlan-VLN (Ours)} & \textbf{Concise (Region-based)} & \textbf{Vectorized Layout} & \Checkmark (RTA/RIA) & \textbf{Linguistic Landmark} & \textbf{Spatio-Semantic Alignment} \\ \bottomrule
\vspace{-1em}
\end{tabular}
}
{\footnotesize
\textit{Abbr: RTA = Region Type Annotations, RIA = Region Identifier Annotations.}
}
\vspace{-2em}
\end{table*}
\subsection{Vision-Language Navigation}
\label{sec:vln}
Vision-Language Navigation (VLN) has advanced significantly in recent years.
Early studies in discrete settings, such as MP3D Simulator~\cite{anderson2018vision}, formulated it as graph traverse task~\cite{fried2018speaker, wang2018look, wang2019reinforced} and mainly focused on grounding instructions with observations~\cite{qi2020object, lin2021scene, moudgil2021soat, gao2021room} or auxiliary information~\cite{an2021neighbor} such as history~\cite{chen2021history} and topological graphs~\cite{chen2021topological, chen2022duet}. 
Substantial progress was achieved through large-scale pre-training~\cite{hao2020prevalent, hong2021vln, Qiao2022HOP, qiao2023hop+} and data augmentation~\cite{he2021landmark, guhur2021airbert, wang2023scalevln, lin2023learning}, enabling agents to approach human performance under this simplified setting.
However, the assumption of a predefined topological graph does not reflect real-world challenges. This has motivated a shift towards continuous environments (VLN-CE)~\cite{krantz_vlnce_2020} in Habitat Simulator~\cite{savva2019habitat}, where agents operate in larger state spaces and must control navigation through low-level actions. 
To bridge the gap between discrete and continuous settings, many works proposed waypoint-based methods~\cite{krantz2021waypoint, hong_2022_bridging_the_gap, an2024etpnav}.

To enhance spatial perception, prior works developed metric and volumetric maps to aggregate multi-modal observations through grid-based memory structures~\cite{wang2023gridmm} and voxelized 3D occupancy representations~\cite{liu2024volumetric}. Recent approaches leveraged pre-trained VLMs to encode structured 3D feature fields or dynamic tokens for cross-modal reasoning~\cite{wang2025g3d, wang2025dynam3d}. 
Besides, 3D Gaussian Maps~\cite{gao20253d} adopted radiance field representations for precise semantic grounding. 
Another direction constructed BEV scene graphs~\cite{an2023bevbert, liu2023bird}, organizing environments into topological graphs with bird’s-eye-view semantic features.

With the development of MLLMs~\cite{gpt3, touvron2023llama, openai2024gpt4,bai2025qwen2, wang2024qwen2}, zero-shot VLN has also gained traction. 
They proposed topological maps~\cite{zhou2024navgpt, chen2024mapgpt, qiao2024opennav} and value maps~\cite{long2024instnav, chen2025canav} for better environment perception and waypoint selection.
More recently, some works collected large-scale navigation videos to finetune Vision-Language-Action models~\cite{zheng2024towards, zhang2025mapnav,zhang2024navid, zhang2024uni, cheng2024navila, wang2025trackvla, wei2025streamvln}, which use pre-trained MLLMs to output low-level actions directly, allowing navigation based solely on RGB observations.

While existing environment representations have significantly advanced spatial awareness, they primarily rely on on-the-fly construction, which restricts the agent to incremental local perception and precludes the use of global spatial priors. 
Attempting to derive a global context through these representations, whether using dense maps~\cite{wang2025g3d, gao20253d, chen2025canav} or structured graphs~\cite{an2024etpnav, chen2022duet}, requires an exhaustive pre-scanning of the entire environment that is both computationally expensive and practically burdensome.
In contrast, our \textbf{FloorPlan-VLN} uses pre-existing floor plans as a low-cost prior as shown in Table~\ref{tab:paradigm_comparison}. 
This approach avoids the inefficient "scan-and-reconstruct" cycle, instead providing a standardized spatial interface for immediate global layout. 
By offloading spatial discovery to this prior, our method enables agents to interpret concise instructions rather than following step-by-step guidance.

\vspace{-5pt}
\subsection{Hand-Drawn Maps for Navigation}
\label{sec:hand-drawn}
\vspace{-4pt}
While the use of global spatial priors has not been considered in VLN, a parallel line of research has investigated navigation with hand-drawn maps.
These studies aim to guide agents in unseen environments with hand-drawn sketches, which provide both the spatial layout and either a reference trajectory or a specified goal.
Early works perceive sketches as an interface for human-robot interaction and focused on extracting navigation clues from sketches based on ruled-based methods~\cite{skubic2002, Parekh2007}. 
Other works explored probabilistic methods such as Monte Carlo Localization~\cite{Matsuo2012, Behzadian2015, Boniardi2015, Boniardi2016}, Hidden Markov Models~\cite{Skubic2003, Skubic2004} and Bayesian Filtering~\cite{xu2024robot}.
Recently, HAM-Nav~\cite{tan2025mobile} built a zero-shot system based on VLM to associate environmental features with hand-drawn maps.

However, such works differ fundamentally from FloorPlan-VLN as they focus on interpreting navigation clues from signals drawn by users rather than following linguistic instructions as illustrated in Table~\ref{tab:paradigm_comparison}. 
Furthermore, the stylistic variance of hand-drawn maps poses a significant domain gap and hinders scalable model training, leading to current research~\cite{tan2025mobile} limited to zero-shot settings.
We argue that standardized floor plans serve as a more robust and scalable \textbf{Unified Spatial Interface}. 
Unlike informal sketches that suffer from high variance, our vectorized representation (i.e., structured region polygons) abstracts the environment layout and region semantics into a consistent format. 
This not only enables scalable model training but also provides a unified target representation that both accurate architectural buleprints and hand-drawn maps can be mapped onto.

\subsection{Floor Plan Point Navigation}
\vspace{-4pt}
\label{sec:floor_plan_nav}
Another line of research explored visual navigation using simplified floor plans.
Early studies focused on floor plan based localization~\cite{wang2019glfp, li2020online, howard2021lalaloc, howard2022lalaloc++, min2022laser, chen2024f3loc}, assuming idealized floor plans containing only wall boundaries.
These approaches often estimated the agent’s location from RGB-D observations or embedded visible points onto the floor plan, followed by querying image features for localization.
More recently, floor plans have been introduced into point-goal navigation tasks~\cite{li2025flona, shen2025vf}, where the goal position is explicitly marked on the map.  
For example, Shen et al.~\cite{shen2025vf} encoded panoramic image features at sampled points on the floor plan and performed localization via a score map, planning trajectories based on a topological graph. 
Li et al.~\cite{li2025flona} instead designed a diffusion policy to generate trajectories and projected them from the floor plan to the real world using the provided scale and offset.

However, these studies typically treat floor plans as oversimplified geometric constraints (e.g., wall boundaries) for localization or point navigation, where the goal is defined by coordinates rather than instructions, as presented in Table~\ref{tab:paradigm_comparison}. 
In contrast, FloorPlan-VLN introduces region-level semantics paired with concise instructions, bridging the gap between low-level geometric navigation and high-level semantic reasoning.
Therefore, our task focuses on the challenge of aligning layout priors, observations, and instructions.

\section{Task and Dataset}
\label{sec:dataset_design}
\subsection{Task Definition}
\vspace{-2pt}
\label{sec: task_setting}
We define the proposed \textbf{FloorPlan-VLN} task as follows. 
Given an unseen environment $\mathcal{E}$, the agent starts from $p_s$ and must reach a goal position $p_g$ following an instruction $I$ while leveraging a floor plan $\mathcal{F}$. 
At each step $t$, it receives an egocentric observation $\mathbf{O}_t$ then predicts an action $a_t \in \mathcal{A}$, where $\small \mathcal{A} = \{\texttt{MoveForward}(d), \texttt{TurnLeft}(\theta), \texttt{TurnRight}(\theta), \texttt{Stop}\}$, $d$ and $\theta$ are fixed step sizes. 
The floor plan $\mathcal{F} = \{ \mathcal{R}_i \}_{i=1}^N$ provides a coarse global layout, containing regions on the same floor (e.g. $N$ regions). 
Each region is represented by a boundary polygon $\mathcal{P}$, a semantic type and a unique identifier.
The boundary polygon is defined by $m$ vertices, i.e., $\mathcal{R}_i = (\mathcal{P}, \text{type}, \text{id}), \mathcal{P} = \{\mathbf{v}_{i} \}_{i=1}^{m}, \mathbf{v}_{i} \in \mathbb{R}^2$.
The instruction $I$ specifies the start region $(\text{id}_s, \text{type}_s)$, goal region $(\text{id}_g, \text{type}_g)$, and stopping conditions. 
Unlike conventional VLN instructions describing landmarks and step-by-step actions, ours are concise and high-level, encouraging reliance on floor plan reasoning.
An episode succeeds if the final position $p_T$ is within a distance threshold $\delta$ of $p_g$.

\begin{figure*}[t]
    \centering
    \includegraphics[width=\linewidth, trim=0 900 0 0, clip]{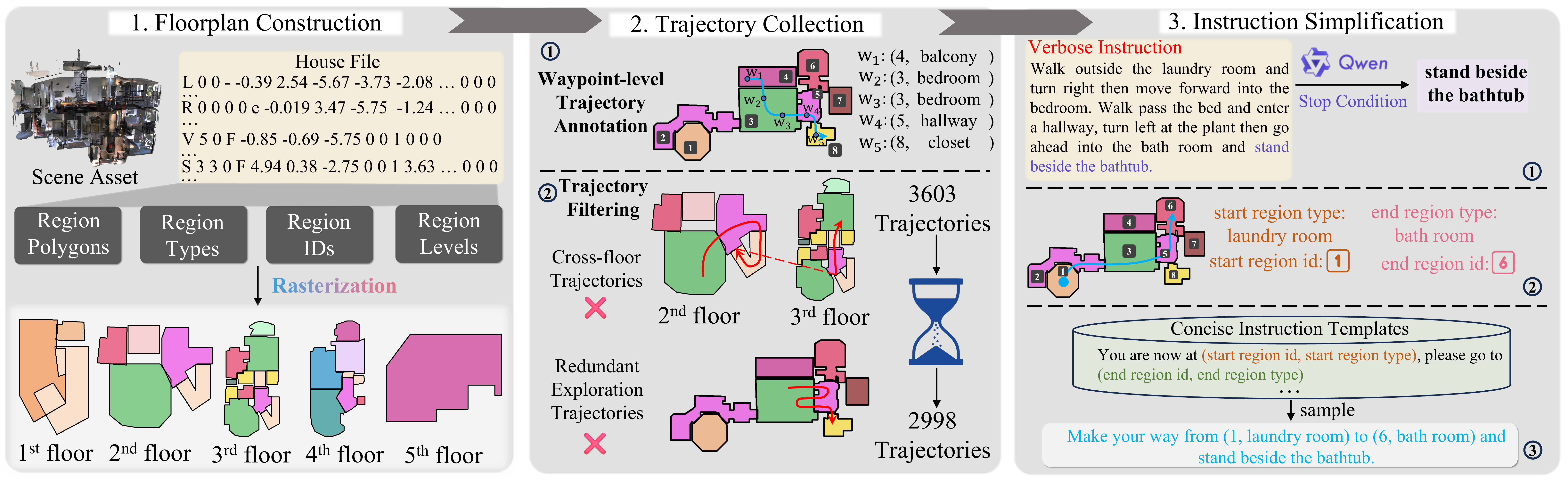}
    \vspace{-2em}
    \caption{The dataset collection pipeline of the FloorPlan-VLN dataset.}
    \label{fig:dataset_collection}
    \vspace{-1em}
\end{figure*}
\vspace{-2pt}

\begin{table*}[t]
\centering
\caption{The \textbf{FloorPlan-VLN Dataset} derives trajectories from two major VLN datasets, R2R and RxR, both built on Matterport3D scenes.
RxR includes longer trajectories and poses greater navigation challenges than R2R. Abbreviations: Val = Validation, \textbf{Trajs} = Trajectories, \textbf{Eps} = Episodes, 
\textbf{FPs} = Floor Plans, \textbf{RTs} = Region Types, \textbf{MTL} = Mean Trajectory Length (meters).
}
\label{tab:dataset_stats}
\vspace{-1em}
\resizebox{\textwidth}{!}{
\begin{tabular}{C{1.5cm}C{1.5cm}C{1.0cm}C{1.0cm}C{1.0cm}C{1.0cm}C{1.0cm}C{1.0cm}|
C{1.5cm}C{1.0cm}C{1.0cm}C{1.0cm}C{1.0cm}C{1.0cm}C{1.0cm}}
\toprule
\multirow{2}{*}{\textbf{Split}} & \multicolumn{7}{c}{\textbf{FloorPlan-R2R Dataset}} & \multicolumn{7}{c}{\textbf{FloorPlan-RxR Dataset}} \\
\cmidrule(lr){2-8} \cmidrule(lr){9-15}
& \textbf{Scenes} & \textbf{Trajs} & \textbf{Eps} & \textbf{FPs} & \textbf{Regions} & \textbf{RTs} & \textbf{MTL} &
\textbf{Scenes} & \textbf{Trajs} & \textbf{Eps} & \textbf{FPs} & \textbf{Regions} & \textbf{RTs} & \textbf{MTL}\\
\midrule
Train & 61 & 2,998 & 9,002 & 111 & 945 & 30  & 9.27 & 59 & 5328 & 15978 & 115 & 942 & 30 & 15.04\\
Val-Seen    & 51 &   226 &   679 &  80 &  885 & 30  & 9.57 & 56 & 593 & 1778 & 94 & 864 & 29 & 14.90\\
Val-Unseen  & 11 &   487 & 1,461 &  18 &  161 & 22  & 9.04 & 11 & 925 & 2775 & 21 & 157 & 22 & 13.97\\
\bottomrule
\end{tabular}}
\vspace{-10pt}
\end{table*}

\begin{figure}[t]
    \centering
    \includegraphics[width=1.0\linewidth, trim=0 110 0 150, clip]{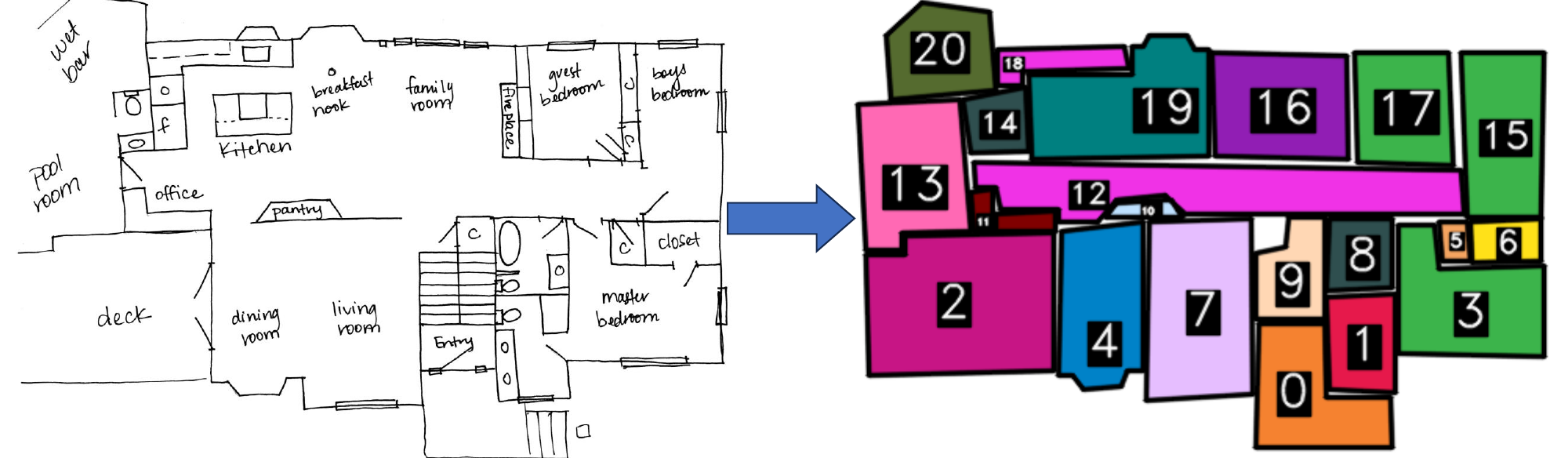}
    \vspace{-2.em}
    \caption{Hand drawn maps can be converted to standardized representation.}
    \label{fig:hand2floor}
    \vspace{-1em}
\end{figure}

\subsection{FloorPlan-VLN Dataset Collection}
\label{sec:dataset-collection}
Unlike hand-drawn maps~\cite{Skubic2004, xu2024robot, tan2025mobile} that vary in style and accuracy, or oversimplified floor plans that contain only wall boundaries, we collect standardized vectorized floor plans.
Architectural floor plans follow inconsistent drafting conventions and often include decorative elements, resulting in non-standardized representations.
To achieve a consistent representation, prior works in floor plan reconstruction~\cite{liu2017raster, yamasaki2018apartment, zeng2019deep, lv2021residential} vectorize raw floor plans into 2D polygons with semantic labels, while recent methods~\cite{yue2023connecting, liu2024polyroom} generate such representations directly from 3D point clouds.
These approaches establish a unified format that encodes spatial structures and region categories.
Following this paradigm, we design a dataset collection pipeline as illustrated in Figure~\ref{fig:dataset_collection}.

Firstly, we construct vectorized floor plans from the Matterport3D dataset~\cite{chang2017matterport3d} by extracting all regions from each floor of every building. 
Each region is represented as a closed polygon defined by its boundary vertexes, annotated with a semantic type, and assigned a unique identifier to distinguish regions of the same type within the same floor. 

\begin{figure}[t]
    \centering
    \includegraphics[width=1.0\linewidth, trim=0 350 0 0, clip]{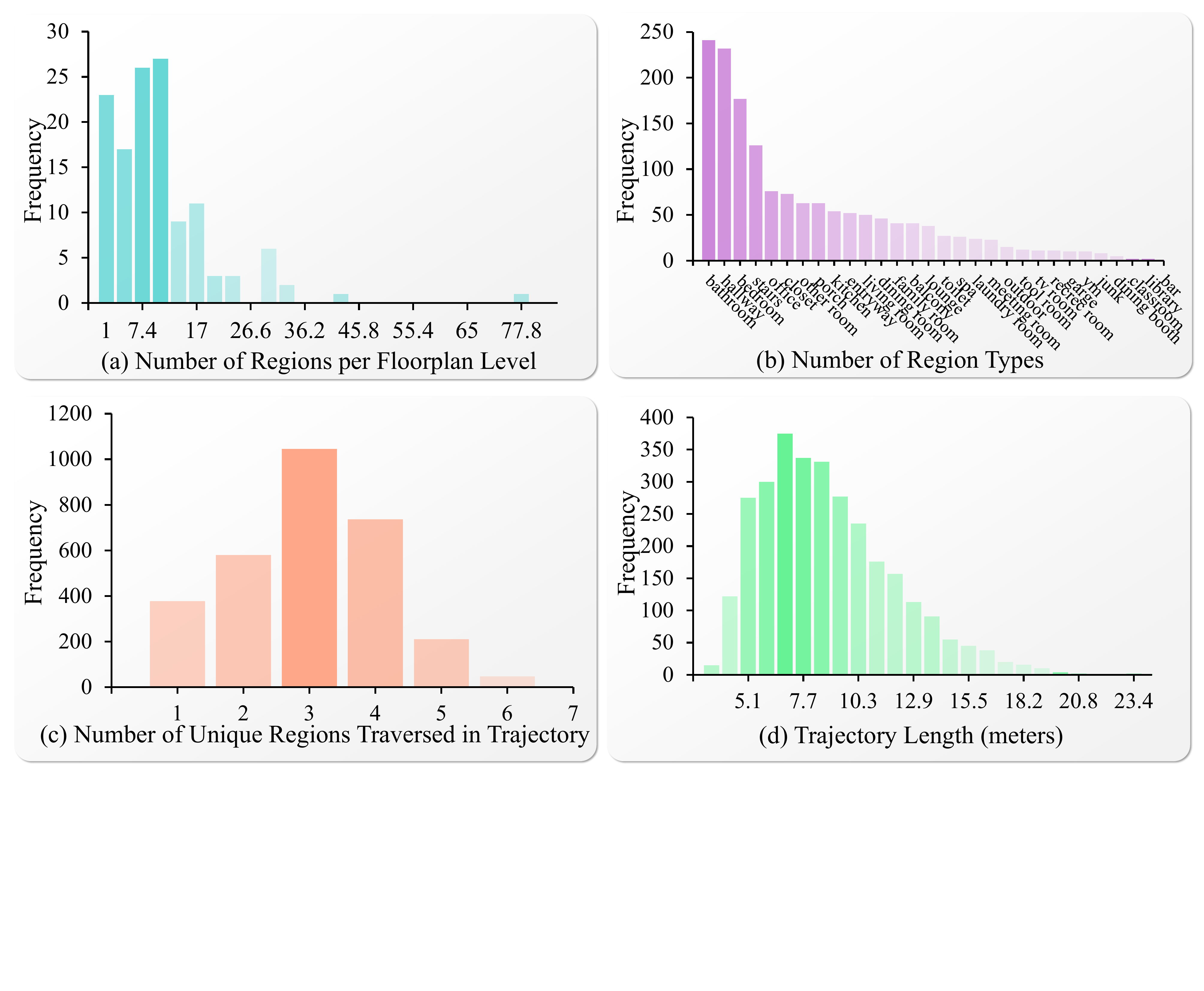}
    \vspace{-2em}
    \caption{Dataset analysis of the FloorPlan-R2R training split.}
    \label{fig:dataset_analysis}
    \vspace{-1em}
\end{figure}
Secondly, we obtain navigation trajectories from R2R-CE and RxR-CE~\cite{krantz_vlnce_2020}. 
For each trajectory, waypoints are mapped to regions by checking which polygon contains them, thereby yielding the sequence of regions traversed by the agent.
We excluding cross-floor and redundant exploration trajectories so that the task focuses on single-floor navigation. 
We focus on single-floor navigation for two main reasons. 
First, statistical analysis of R2R-CE trajectories shows that cross-floor episodes account for only 2.2\% of the dataset, lacking enough supervision for learning vertical transitions and floor plan switching. 
Second, the core challenge of FloorPlan-VLN is spatio-semantic alignment between global layout priors and egocentric observations. 
Although multi-floor navigation is feasible based on staircase detection and dynamic floor plan switching, it would shift focus from cross-modal reasoning. 
We argue that robust single-floor navigation is a prerequisite for future multi-floor extensions.

Thirdly, to reduce the redundancy of the original instructions and encourage the use of floor plans, we simplify them into concise forms using Qwen-2.5-VL~\cite{wang2024qwen2}. 
For each episode, the concise instructions specify only the start and target regions, along with their corresponding identifiers and semantic types. 
Meanwhile, the stopping conditions are extracted directly from the original instructions to preserve authentic human intent.
We design a set of 10 concise instruction templates that describe the task as shown in Figure~\ref{fig:dataset_collection}.
This procedure retains essential task requirements while discarding detailed descriptions of landmarks and step-by-step actions. 

To ensure practical applicability, we develop a semi-automated pipeline that can efficiently convert hand-drawn maps into our standardized format as demonstrated in Figure~\ref{fig:hand2floor}.
This pipeline is also used in our real-world experiments, and more details will be introduced in \S~\ref{sec:real-world-implementation-details}.

\begin{figure*}[t]
    \centering
    \includegraphics[width=\linewidth, trim=0 600 100 100, clip]{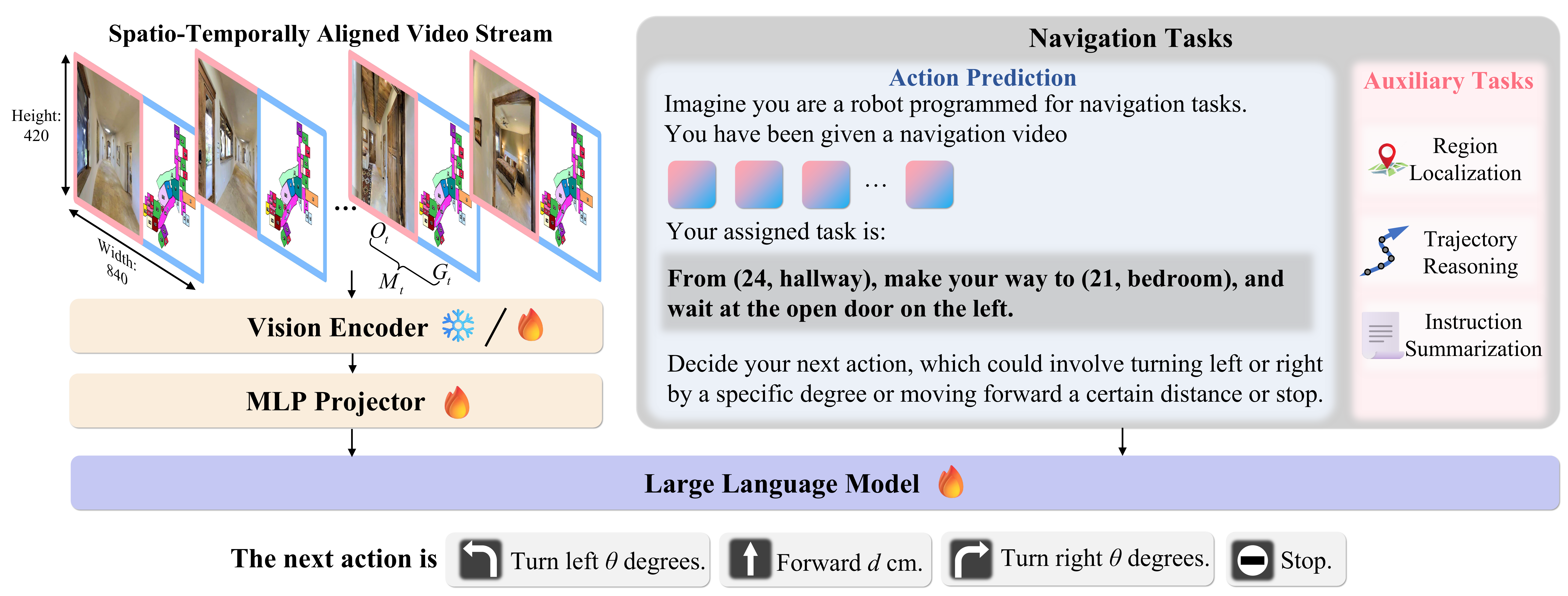}
    \vspace{-2em}
    \caption{Overview of the FP-Nav framework.
The model takes a spatio-temporally aligned dual-view video stream as input, jointly encoding floor plans and egocentric observations. Three auxiliary tasks are designed to strengthen cross-modal alignment, and the model autoregressively predicts navigation actions.}
\label{fig:framework}
\vspace{-1.5em}
\end{figure*}
\vspace{-4pt}
\subsection{Dataset Analysis}
\vspace{-2pt}
As shown in Table~\ref{tab:dataset_stats}, the FloorPlan-VLN dataset unifies trajectories from R2R and RxR.
The R2R split provides shorter, simpler trajectories, while RxR includes longer and more complex trajectories, posing greater challenges.
Each dataset is divided into \textit{train}, \textit{val-seen}, and \textit{val-unseen} splits, where the seen split tests generalization to new trajectories within familiar scenes, and the unseen split evaluates generalization to novel environments and floor plans.

In total, FloorPlan-VLN spans 70+ buildings, 130+ floor plans, and over 1k region-level annotations across 30 semantic region types, offering diverse spatial structures and scene scales for studying floor plan guided navigation.
Moreover, it introduces vectorized region polygons enriched with region types, identifiers, and instructions, establishing the first benchmark for studying navigation under floor plan priors with concise instructions.
To further characterize the dataset, we analyze the training split of FloorPlan-R2R from four perspectives, as illustrated in Figure~\ref{fig:dataset_analysis}.
Figure~(a) shows that most floor plans contain more than 7 regions, indicating that the dataset includes environments of varying scales and spatial complexity.
Figure~(b) presents the distribution of region types.
Figure~(c) shows that most trajectories traverse at least three regions, and Figure~(d) reveals that their average length is about 9 meters, highlighting the long-horizon and multi-region challenges of the FloorPlan-VLN task.

\section{FP-Nav}
\vspace{-2pt}
\label{sec:Method}
MLLMs pretrained on web-scale data implicitly learn to interpret spatial diagrams such as floor plans.
To exploit this ability for navigation, we propose FP-Nav, which rasterizes floor plans as images and finetunes Qwen-2.5-VL~\cite{bai2025qwen2} to align them with egocentric observations and instructions for action generation.
As illustrated in Figure~\ref{fig:framework}, FP-Nav comprises a vision encoder for extracting visual features, a lightweight MLP projector for vision-language feature alignment, and an LLM that performs cross-modal fusion and generates low-level navigation actions.
Given an instruction, a sequence of egocentric observations, and a floor plan, we finetune the model to jointly align these modalities and generate step-wise actions.
To address the alignment challenge between floor plan and observation, we introduce a spatio-temporally aligned video pairing strategy.
Besides, three auxiliary tasks are designed to strengthen FP-Nav's capabilities in region localization, trajectory reasoning and instruction summarization.

\subsection{Spatio-Temporally Aligned Video Stream}
\begin{figure}[t]
    \centering
    \includegraphics[width=1.0\linewidth, trim=0 20 0 0, clip]{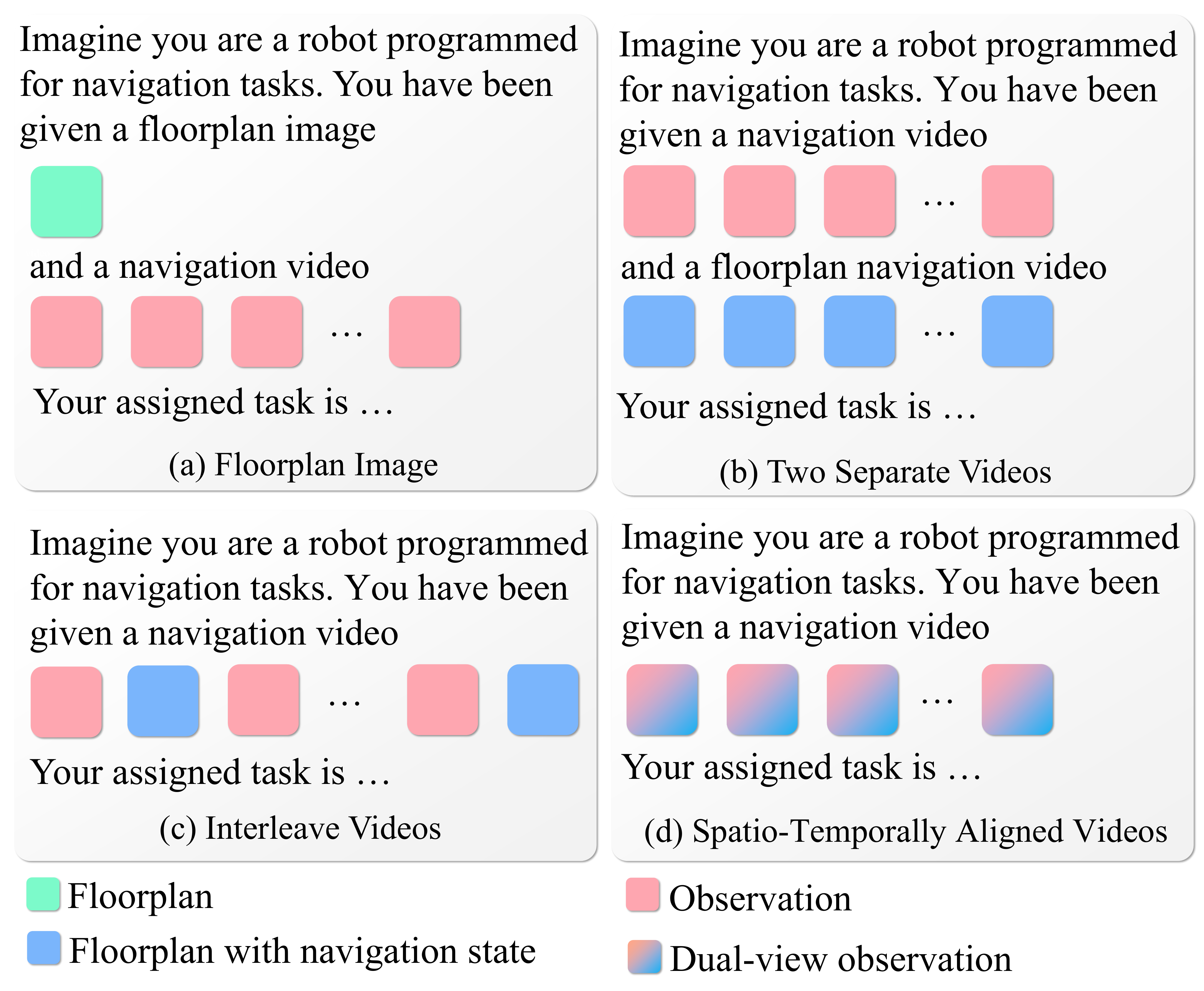}
    \vspace{-2em}
    \caption{Comparison of four input strategies.}
    \label{fig:input_strategies}
    \vspace{-1em}
\end{figure}

The collected floor plans are vectorized as polygons which are a set of 2D point clouds. A straightforward method to encode them is using point encoders~\cite{qi2017pointnet, qi2017pointnet++} to extract their geometry features then concatenate them with corresponding semantic type features and ID features.
However, we empirically find that training a floor plan encoder to align its learned features with the MLLM's feature space is difficult, since MLLMs are primarily trained on rasterized visual data such as images.
Thus we decide to rasterize floor plans $\mathcal{F}$ into images $\mathbf{I}_{\mathcal{F}} \in \mathbb{R}^{H \times W \times 3}$, where $H$ and $W$ represent image's height and width, respectively.
To embed the semantic structure into the image, we employ visual prompting~\cite{yang2023set}: Region Types (e.g., living room, kitchen) are encoded by distinct colors for easy visual segmentation, and unique region identifiers are marked with numeric labels.
To establish effective correspondences between observations and floor plans, we explore four input strategies that progressively transition from implicit to explicit alignment as shown in Figure~\ref{fig:input_strategies}.

\textbf{(a) Static Separate Input.} The most straightforward method provides the static floor plan and the egocentric observation sequence as separate inputs as shown in Figure~\ref{fig:input_strategies} (a). 
It requires the agent to implicitly localize itself according to a static floor plan and dynamic observations which remains a significant open challenge.
As the first exploration, we aim to prioritize the core challenge of aligning egocentric observations with floor plans and instructions.
Therefore, we utilize the agent's ground-truth poses from simulator as oracle signals and design additional experiments to evaluate the robustness of our method to pose perturbations and floor plan distortions.
This setup establishes the first performance upper bound and decouples the cross-modal alignment challenge from the complexities of location estimation. 

\textbf{(b) Dual-Stream Temporal Fusion.} Utilizing the oracle pose, we generate a dynamic floor plan sequence where the agent's historical trajectory and current pose are explicitly rendered (blue squares).
In this approach, the observation video and the dynamic floor plan sequence are fed as two independent streams, forcing the model to learn cross-modal correspondences across the entire temporal context.

\textbf{(c) Interleaved Frame Fusion.} To facilitate more granular alignment at each time step, we interleave the egocentric frames and their corresponding allocentric floor plan frames. 
This strategy encourages the model to perform alternating cross-modal comparisons within the visual sequence, trying to reduce the difficulty of long-range temporal matching.

\textbf{(d) Spatio-Temporally Aligned Dual-View.} Inspired by how humans use navigation tools, where they simultaneously observe their surroundings and their pose on the map, we propose a spatio-temporally aligned video stream as input as shown in Figure~\ref{fig:input_strategies} (d).
This strategy shifts the learning burden from implicit alignment to explicit spatial correlation.
At each time step $t$, the dual-view input frame $\mathbf{M}_t$ is constructed by horizontally concatenating the egocentric observation $\mathbf{O}_t$ with the floor plan image $\mathbf{G}_t$, where $\mathbf{G}_t$ includes the agent's historical trajectory $\mathbf{T}_{0:t-1} = {s_0, s_1, \dots, s_{t-1}}$ and current pose $\mathbf{s}_t = (x_t, y_t, \theta_t)$ explicitly marked:
\begin{align}
\small
\mathbf{M}_t = [\mathbf{O}_t \parallel \mathbf{G}_t]; \ \mathbf{G}_t = \mathcal{R}(\mathbf{I}_{\mathcal{F}}, \mathbf{T}_{0:t-1}, \mathbf{s}_t)
\end{align}
where $\mathcal{R}(\cdot)$ is the rendering function for floor plan $\mathbf{I}_{\mathcal{F}}$ given the trajectory $\mathbf{T}_{0:t-1}$ and pose $\mathbf{s}_t$, and $[\cdot \parallel \cdot]$ denotes the horizontal concatenation function.
This provides the visual encoder with an immediate spatial-temporal correspondence.

We denote the navigation task's dataset as $\mathcal{D}_{\text{nav}}=\{(\mathbf{Q}_1, \mathbf{A}_1), (\mathbf{Q}_2, \mathbf{A}_2), \cdots, (\mathbf{Q}_N, \mathbf{A}_N)\},\ \text{where }\mathbf{Q}_t=\{\text{prompts}; \mathbf{M}_{t-h:t};I\}$ (as shown in Figure~\ref{fig:framework}) where $h$ is the length of video sequence and $\mathbf{A}_t=\{\text{The next action is }a_t.\}$

\begin{figure}[t]
    \centering
    \includegraphics[width=1.0\linewidth, trim=0 0 0 0, clip]{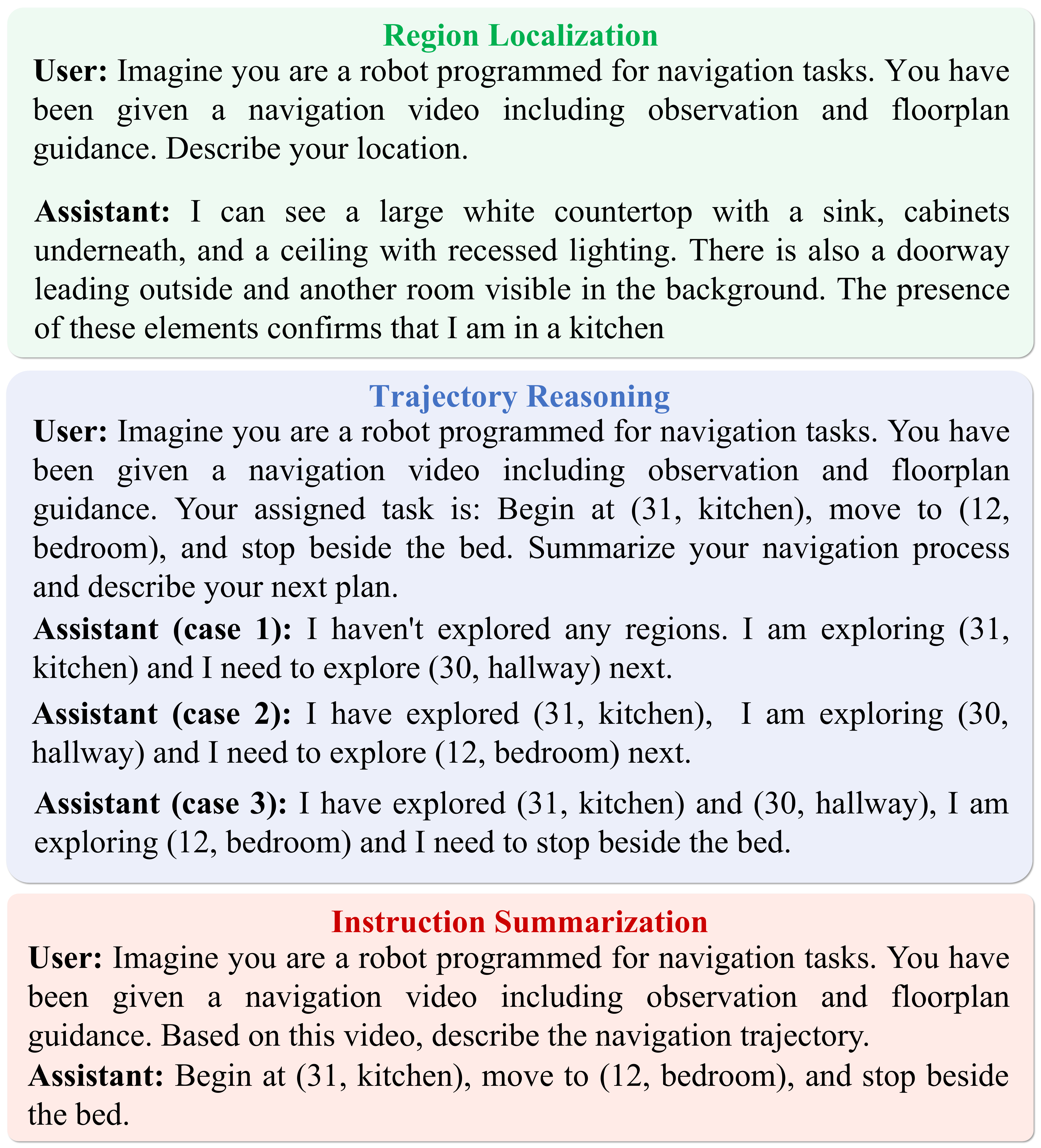}
    \vspace{-2em}
    \caption{Three auxiliary tasks designed to strengthen the alignment among instructions, egocentric observations, and floor plans.}
    \label{fig:auxiliary_tasks}
    \vspace{-1em}
\end{figure}

\subsection{Auxiliary Tasks}
\vspace{-3pt}
\label{sec:auxiliary-tasks}
To strengthen the model’s spatio-semantic alignment and high-level reasoning capabilities, we incorporate three auxiliary tasks alongside the primary navigation objective.
These tasks encourage the MLLM to internalize the relationship between egocentric observations, floor plans, and instructions.

\textbf{Region Localization.}
This task encourages the agent to associate current visual perceptions with the floor plan. 
At randomly sampled steps $t$ within a trajectory, the agent is prompted to describe its observation $\mathbf{O}_t$ and infer the current region type $\mathbf{R}_t$.
We employ a teacher model (e.g., Qwen-2.5-VL-32B) to generate a concise caption $\mathbf{C}_t$ for the frame. 
The ground-truth region type $\mathbf{R}_t$ is automatically determined by a spatial query $f_{\text{map}}(\mathbf{s}_t)$ that checks which polygonal region boundary in the vectorized floor plan contains the agent’s current pose $\mathbf{s}_t$.
The task's QA dataset is similar to $\mathcal{D}_{\text{nav}}$ while the target response $\mathbf{A}_t=\{\mathbf{C}_t;\mathbf{R}_t\}$ as shown in Figure~\ref{fig:auxiliary_tasks}.

\textbf{Trajectory Reasoning.}
To enhance long-term spatial memory and progress tracking, the agent is asked to summarize its navigation history, report its current progress, and generate a plan for the next step.
Each step of the trajectory is annotated with the agent’s current region identifier and region type, as introduced in \S~\ref{sec:dataset-collection}.
These annotations allow us to construct ground truth supervision by tracing agent's visited region sequence and identify the next region to traverse.
As illustrated in Figure~\ref{fig:auxiliary_tasks}, we consider three cases:
\begin{itemize}
    \item \textbf{Initialization Stage.} When the agent has not yet explored any regions, it focuses on identifying its current region $\mathbf{R}_t$ and the next one to visit $\mathbf{P}_t$.
    \item \textbf{Navigation Stage.} As the agent explores multiple regions, it is required to summarize the visited regions in order $\mathbf{H}_t$, identify its current region and type $\mathbf{R}_t$, and predict the next region to explore $\mathbf{P}_t$.
    \item \textbf{Termination stage.} Once the agent reaches the region containing the goal, it summarizes all previous explorations, infers its current location, and describes the stop condition $\mathbf{S}_t$ specified in the instruction.
\end{itemize}

The dataset $\mathcal{D}_{traj}$ is similar to Region Localization task while the target response $\textbf{A}_t$ consists of $\{\mathbf{H}_t,\mathbf{R}_t, \mathbf{P}_t(\mathbf{S}_t)\}$.

\textbf{Instruction Summarization.}
This task reinforces the alignment between instruction and spatio-temporally visual inputs. 
Given a successful dual-view navigation video $\mathbf{M}_{1:T}$, the agent is required to reconstruct the original concise instruction $I$ (encompassing the start, goal, and stop condition).
By learning to reconstruct the high-level intent behind a navigation episode, the model develops a deeper understanding of how concise instructions correspond to observations and floor plans.
This dataset is denoted as $\mathcal{D}_{\text{sum}}$ with its inputs $\mathbf{Q}_t=\{\mathbf{M}_{1:T}\}$ and target response $\mathbf{A}_t=\{I\}$.

\vspace{-4pt}
\subsection{Unified Training Objective}
\label{sec:unified training objective}
While these tasks are functionally diverse, we formulate the entire training process as a unified Next Token Prediction (NTP) problem. 
This approach treats the navigation task and auxiliary tasks as different parts of the same autoregressive model, distinguished by specific text prompts.
We aggregate all tasks into a unified dataset $\mathcal{D} = \mathcal{D}_{\text{nav}} \cup \mathcal{D}_{\text{aux}}$, where $\mathcal{D}_{\text{aux}} = \mathcal{D}_{\text{loc}} \cup \mathcal{D}_{\text{traj}} \cup \mathcal{D}_{\text{sum}}$. 
For any sample in $\mathcal{D}$, the input consists of a multi-modal sequence $\mathbf{X}=\{\mathbf{M};\text{prompts}, I\}$, and the target is a sequence of tokens $\mathcal{Y} = \{y_1, y_2, \dots, y_N\}$.
The model is trained by minimizing the unified cross-entropy loss:$$\mathcal{L}_{\text{total}} = - \sum_{(\mathbf{X}, \mathcal{Y}) \in \mathcal{D}} \sum_{i=1}^{N} \log P(y_i \mid \mathbf{X}, y_{<i}; \theta)$$where $\theta$ represents the trainable parameters. 
During training, we employ a task-balanced sampling strategy to ensure that the primary navigation task $\mathcal{D}_{nav}$ remains the dominant objective while benefiting from the cross-modal alignment provided by $\mathcal{D}_{\text{aux}}$.
More details are introduced in \S~\ref{sec:implementation details}
This unified formulation allows the MLLM to naturally share internal representations across execution and reasoning.

\subsection{Modeling Actuation and Floor Plan Noises}
\label{sec:noise}
To evaluate the system's robustness in non-ideal conditions, we introduce a comprehensive uncertainty framework that simulates real-world control errors and prior inaccuracies.

\textbf{Actuation Noises:} In real-world deployment, the robot controller could be imperfect, therefore we model the actuation noise by perturbing the agent's actual displacement.
Given an action $a_t \in \{\texttt{MoveForward},\texttt{TurnLeft}, \texttt{TurnRight}\}$, the agent's pose $\mathbf{s}_t=(x_t, y_t, \theta_t)$ is updated via a noisy transition model.
For rotation:
\begin{align}
\small
\theta_{t+1} = \theta_t + \text{sgn}(a_t) \cdot \Delta\hat{\theta} + \epsilon_{\text{rot}}, \quad \epsilon_{\text{rot}} \sim \mathcal{N}(0, \sigma_{\text{rot}}^2)
\end{align}
where  $\Delta\hat{\theta}$ is the nominal turn angle and $\epsilon_{\text{rot}}$ represents the heading noise.

For moving forward: 
\begin{equation}
\begin{aligned}
&\left\{
\begin{aligned}
d_t &= \hat{d}(1 + \epsilon_m), 
& \epsilon_m &\sim \mathcal{N}(0, \sigma_{\text{move}}^2) \\
\theta_{t+1} &= \theta_t + \epsilon_d, 
& \epsilon_d &\sim \mathcal{N}(0, \sigma_{\text{drift}}^2)
\end{aligned}
\right.
\\[1em]
&\left\{
\begin{aligned}
x_{t+1} &= x_t + d_t \sin(\theta_{t+1}) \\
y_{t+1} &= y_t + d_t \cos(\theta_{t+1})
\end{aligned}
\right.
\end{aligned}
\end{equation}

where $\hat{d}$ is the nominal step size, $\epsilon_m$ is the relative distance error, and $\epsilon_d$ accounts for the subtle heading drift during straight-line motion.
We set $\sigma_{\text{drift}}=0.1\sigma_{\text{move}}$

\textbf{Floor Plan Noises}: To simulate the inaccuracies in real-world floor plans and hand-drawn maps, we introduce two types of geometric noises. 
Rather than assuming a perfect map-to-world alignment, these noises evaluate the agent's spatial reasoning robustness under imperfect global priors.

\textit{Global Scale Variance}: In practical deployment, floor plans or hand-drawn maps may contain scale inaccuracies relative to the real environment. 
As a result, the agent’s trajectory projected onto the map may appear to be deviated.
To equivalently simulate the consequences of this scaling discrepancy, we inject a global scale multiplier $\alpha$ into the trajectory projection process, rather than altering the map itself:
\begin{equation}
\small
\hat{\mathbf{p}}_t = \alpha \cdot f_{\text{proj}}(\mathbf{p}_t), \quad \alpha \sim \mathcal{N}(1, \sigma_{\text{scale}}^2)
\end{equation}
where $\mathbf{p}_t = (x_t, y_t)$ is the agent's actual position in the simulator, $f_{\text{proj}}(\cdot)$ is the projection function, and $\hat{\mathbf{p}}_t$ is the corresponding coordinate rendered on the floor plan image. 

\textit{Geometric Jitter}: Imperfect floor plans and hand-drawn maps may include shape distortions.
To emulate this real-world variance, we apply a random displacement to each vertex $\mathbf{v}_i \in \mathcal{P}$ of the region boundaries:
\begin{equation}
\small
v'_i = v_i + \mathbf{\epsilon}_i,\mathbf{\epsilon}_i \sim \mathcal{N}(0, \sigma_{\text{jitter}}^2)
\end{equation}
The primary objective of introducing this perturbation is to explicitly verify that FP-Nav relies on a holistic understanding of global spatial layout for navigation, rather than overfitting to rigid pixel-level coordinate dependencies.

\section{Experiment}
\label{sec:experiment}
\vspace{-3pt}
\begin{table*}[t]
\centering
\caption{Comparison with baselines across FloorPlan-R2R and FloorPlan-RxR.}
\vspace{-0.75em}
\label{tab:main_results}
\resizebox{\textwidth}{!}{
\begin{tabular}{
C{2.5cm} |
C{0.55cm} C{0.55cm} C{0.55cm} C{0.6cm} |
C{0.55cm} C{0.55cm} C{0.55cm} C{0.6cm} |
C{0.55cm} C{0.55cm} C{0.55cm} C{0.6cm} |
C{0.55cm} C{0.55cm} C{0.55cm} C{0.6cm}
}
\toprule
\multirow{3}{*}{\textbf{Method}} 
& \multicolumn{8}{c|}{\textbf{FloorPlan-R2R}} 
& \multicolumn{8}{c}{\textbf{FloorPlan-RxR}} \\
\cmidrule(lr){2-9}\cmidrule(lr){10-17}
& \multicolumn{4}{c|}{\textbf{Val Seen}} 
& \multicolumn{4}{c|}{\textbf{Val Unseen}} 
& \multicolumn{4}{c|}{\textbf{Val Seen}} 
& \multicolumn{4}{c}{\textbf{Val Unseen}} \\
\cmidrule(lr){2-5}\cmidrule(lr){6-9}\cmidrule(lr){10-13}\cmidrule(lr){14-17}
& NE$\downarrow$ & OSR$\uparrow$ & \textbf{SR}$\uparrow$ & \textbf{SPL}$\uparrow$
& NE$\downarrow$ & OSR$\uparrow$ & \textbf{SR}$\uparrow$ & \textbf{SPL}$\uparrow$
& NE$\downarrow$ & OSR$\uparrow$ & \textbf{SR}$\uparrow$ & \textbf{SPL}$\uparrow$
& NE$\downarrow$ & OSR$\uparrow$ & \textbf{SR}$\uparrow$ & \textbf{SPL}$\uparrow$ \\
\midrule
Qwen-zs              & 9.6 & 3.1  & 2.2  & 1.9  & 9.3 & 3.4  & 2.4  & 2.0  & 12.0 & 8.6 & 5.6 & 4.4 & 11.1 & 11.2 & 7.2 & 5.7 \\
NaVILA-zs              & 9.9 & 19.0  & 8.8  & 0.6  & 10.1 & 17.6  & 5.5  & 3.5  & 12.9 & 20.1 & 7.1 & 4.5 & 11.2 & 20.0 & 6.9 & 4.4 \\
StreamVLN-zs              & 9.1 & 24.3  & 11.5  & 6.5 & 8.8 & 30.3  & 14.6  & 7.0  & 11.6 & 27.9 & 15.0 & 9.2 & 10.8 & 26.9 & 14.4 & 7.7 \\
InternVLA-N1-zs              & 7.3 & 26.5  & 10.6  & 9.2  & 8.2 & 19.4  & 7.5  & 7.0  & 10.4 & 22.4 & 12.3 & 9.7 & 10.0 & 22.4 & 10.0 & 8.0 \\
Navid-zs             & 9.5 & 3.1  & 2.0  & 1.6  & \textbf{8.9} & 2.9  & 1.8  & 1.0  & 11.6 & 10.8 & 6.7 & 5.0 & 10.7 & 12.1 & 10.5 & 8.4 \\
Navid-ft             & 8.8 & 27.0 & 18.1 & 14.2 & 9.1 & 25.2 & 17.0 & 13.0 & 11.3 & 20.5 & 12.8 & 8.6 & 10.6 & 24.2 & 13.4 & 9.2 \\
\textbf{FP-Nav}      & 8.6 & 36.3 & 23.0 & 17.8 & 9.8 & 35.2 & 20.9 & 15.3 & 11.9 & 26.8 & 17.0 & 12.2 & 11.3 & 28.9 & 15.1 & 10.2 \\
\textbf{FP-Nav-v}    & 8.2 & 48.2 & 38.1 & 31.5 & 9.6 & 37.4 & 25.9 & 21.7 & 11.9 & 29.8 & 18.7 & 15.0 & 11.3 & 31.2 & 17.7 & 14.1 \\
\textbf{FP-Nav-v-rxr}& \textbf{7.0} & \textbf{51.1} & \textbf{43.2} & \textbf{38.4} & 9.4 & \textbf{39.8} & \textbf{28.8} & \textbf{24.0} & \textbf{9.9} & \textbf{37.3} & \textbf{24.1} & \textbf{20.0} & \textbf{10.1} & \textbf{38.5} & \textbf{25.2} & \textbf{20.1} \\
\bottomrule
\end{tabular}}
\vspace{-1em}
\end{table*}

We design experiments to address the following questions:
(1) How does our method compare with other adapted SoTA baselines?
(2) Does our method robust to imperfect pose and floor plans?
(3) Is the proposed spatio-temporally aligned video stream effective?
(4) Does the model learn to align the floor plan with egocentric observations?
(5) What is the impact of different auxiliary tasks?
(6) What is the effect of training different parameters (e.g., MLP/LLM/vision encoder)?

\subsection{Experiment Setup}
All experiments are conducted in the Habitat simulator~\cite{savva2019habitat}, which provides continuous state in reconstructed indoor environments. 
We evaluate on the validation-seen and validation-unseen splits  
and report: Navigation Error (NE) measures the mean distance from the final location to the destination; Success Rate (SR) represents the proportion of episodes with NE under 3 meters; Oracle Success Rate (OSR) calculates SR with an oracle stop policy and Success weighted by Trajectory Length (SPL) normalizes SR by trajectory length.

\begin{figure}[t]
    \centering
    \includegraphics[width=1.0\linewidth, trim=0 750 0 0, clip]{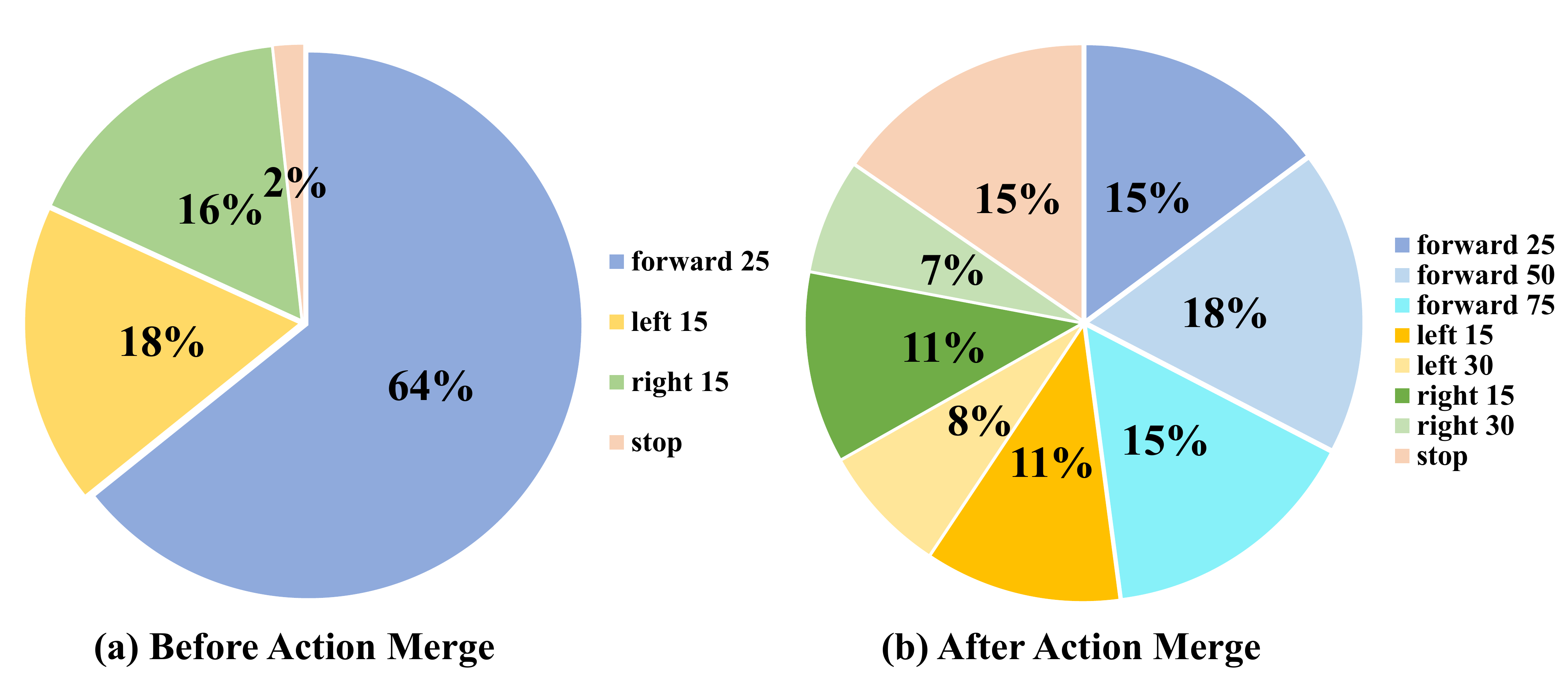}
    \vspace{-2em}
    \caption{Action frequency after merging consecutive steps in the FloorPlan-R2R finetuning dataset.}
    \vspace{-1em}
    \label{fig:r2r_data_distribution}
\end{figure}
\subsection{Implementation Details}
\label{sec:implementation details}
We collect navigation videos for finetuning by executing ground-truth trajectories in the Habitat simulator.
The translation step $d$, rotation step $\theta$ and success distance threshold $\delta$ introduced in Section~\ref{sec: task_setting}, are set to 25 cm, 15$^{\circ}$ and 3 m respectively.
The video sequence length $h$ is set to 6.
During execution, the agent follows each trajectory step by step while capturing egocentric observations at 1 fps.
The recorded pose data are then projected onto the floor plan, depicting the agent’s pose and trajectory over time.

To balance the action distribution, we merge consecutive identical actions (e.g., forwards) and upsample underrepresented ones (e.g., stop).
As shown in Figure~\ref{fig:r2r_data_distribution}(a), the original data distribution is highly imbalanced.
Most actions are Move Forward, while Stop and turning actions occur much less frequently.
Training directly on such a distribution would make it difficult for the model to learn effective turning or stopping behaviors, leading to suboptimal navigation.

To address this, we merge consecutive Move Forward actions (e.g., “move forward 50 cm” and “move forward 75 cm”) into a single longer step action to reduce redundancy. 
Similarly, consecutive Turn Left or Turn Right actions are merged to preserve meaningful rotation patterns.
During inference, we decompose the predicted action into a sequence of primitive actions and execute them sequentially.
For underrepresented actions such as Stop, we upsample them by repetition to increase their frequency during training.
After balancing, the action distribution becomes significantly more uniform, as shown in Figure~\ref{fig:r2r_data_distribution}(b), allowing the model to learn a more stable and comprehensive navigation policy.
In total, we obtain approximately 350K action QA samples. 

For auxiliary tasks, we additionally collect about 174K QA samples for the region localization and trajectory reasoning tasks, and 9K samples for the instruction summarization task.

We finetune FP-Nav using 4 H100 GPUs for 20 hours (80 GPU hours in total).
For navigation episodes with long step sequences, we truncate the video to at most 6 frames, keeping the first and last frames and uniformly sampling the remaining ones.
The batch size is set as 4 and the learning rate is $4\times10^{-5}$. 
During inference, we follow~\cite{zhang2024navid} and use expression matching to parse action commands from the model’s textual output.

\subsection{Main Results}
As FloorPlan-VLN is a new task setting, we establish five zero-shot baselines and one finetuned baseline, and compare them with our proposed FP-Nav on the FloorPlan-VLN benchmark.
The evaluated methods are as follows where ``\textbf{-zs}'' means this method is evaluated in a zero-shot manner and all baselines are evaluated with our dual-view, spatio-temporally aligned input strategy:

\begin{itemize}
\item \textbf{Qwen-zs:} We apply Qwen-2.5-VL-7B for step-wise action generation on both FloorPlan-VLN-R2R and FloorPlan-VLN-RxR datasets.
\item \textbf{NaVILA-zs:} NaVILA~\cite{cheng2024navila} is a Vision-Language-Action (VLA) model pre-trained on VLN-CE videos and large-scale human navigation videos.
\item \textbf{StreamVLN-zs:} StreamVLN~\cite{wei2025streamvln} is a video-based LLM which supports multi-turn dialouge navigation through a hybrid slow-fast context modeling strategy.
\item \textbf{InternVLA-N1-zs:} InternVLA-N1~\cite{internvla-n1} is a dual-system framework which uses MLLM as slow-system for goal point selection and uses a navigation diffusion model~\cite{cai2025navdp} as fast-system for trajectory generation.
\item \textbf{Navid-zs:} Navid~\cite{zhang2024navid} is a VLA model pre-trained on VLN-CE videos that designs grid sampling strategy for better memory efficiency.
\item \textbf{Navid-ft:} Finetunes Navid with FloorPlan-VLN data.
\item \textbf{FP-Nav:} Adopts Qwen-2.5-VL-7B as the backbone while freezing the vision encoder for fair comparison with Navid-ft.
Ablation studies are conducted on this setup.
\item \textbf{FP-Nav-v:} Unfreezes Qwen’s vision encoder, allowing updates to the ViT~\cite{dosovitskiy2020vit} patch embeddings and Rotary Positional Embeddings (RoPE)~\cite{su2024rope}.
\item \textbf{FP-Nav-v-rxr:} Further augments training with approximately 1.7k long-horizon episodes (trajectory length $>$10 m) from the FloorPlan-RxR dataset to improve extended navigation ability.
\end{itemize}

\textbf{Results on FloorPlan-R2R.} As shown in Table~\ref{tab:main_results}, most of the methods achieve relative low zero-shot performance in our new paradigm.
While specialized navigation agents such as StreamVLN achieve a higher zero-shot success rate (SR), we attribute this primarily to their intensive domain-specific pre-training on VLN-CE datasets and their architectures tailored for navigation tasks.
To decouple the effect of pre-existing navigation biases from the actual contribution of floor plan priors, we intentionally select Qwen-2.5-VL as our primary backbone. 
The marginal zero-shot performance of vanilla Qwen (Qwen-zs) underscores a fundamental challenge: general-purpose MLLMs struggle to ground concise instructions within floor plans and egocentric observations without explicit fine-tuning. 
Consequently, utilizing a navigation-agnostic generalist model ensures a more rigorous and objective evaluation. 

Beyond finetuning MLLM, we also finetune Navid to demonstrate the transferability of our proposed spatio-temporally aligned dual-view input strategy.
Among the established baselines, StreamVLN relies on a multi-turn dialogue state tracking architecture, NaVILA benefits from extensive augmentation with large-scale human navigation data, and InternVLA-N1 employs a complex dual-system reasoning framework. 
These specialized designs introduce additional confounding variables that make it difficult to isolate the specific impact of the floor plan prior.
Therefore, we specifically choose to fine-tune Navid.

After finetuning, Navid-ft achieves a significant improvement, demonstrating that the model can partially adapt to floor plan guided navigation. Our FP-Nav, further improves OSR, SR, and SPL by +10.0, +3.9, and +2.3 points on Val-Unseen over Navid-ft.
Although Navid is finetuned on VLN-CE, it is built on LLaMA-Vid~\cite{li2024llama-vid}, whose pretraining focuses on video understanding with limited exposure to schematic or spatial diagram reasoning.
This indicates that MLLMs with stronger spatial and diagram oriented pretraining are inherently better at learning correspondences between floor plans and egocentric observations, even without updating the visual encoder.

Unfreezing Qwen’s vision encoder (FP-Nav-v) brings substantial gains, with SR and SPL improving by 24\% and 42\% on Val-Unseen split, respectively.
This suggests that updating the visual backbone helps the model transition from natural image understanding to the symbol-dense floor plan domain, where ViT patch embeddings and RoPE can adjust to the dual-view temporal alignment, yielding improved cross-view consistency.
Finally, FP-Nav-v-rxr leads to consistent improvements across all metrics, confirming the benefit of longer trajectories for enhancing navigation robustness.
We also observe higher performance on Val-Seen compared to Val-Unseen after unfreezing, suggesting stronger memorization and reasoning over previously seen floor plans, which benefits intra-scene generalization.

\begin{figure}[t]
    \centering
    \includegraphics[width=1.0\linewidth, trim=0 600 0 0, clip]{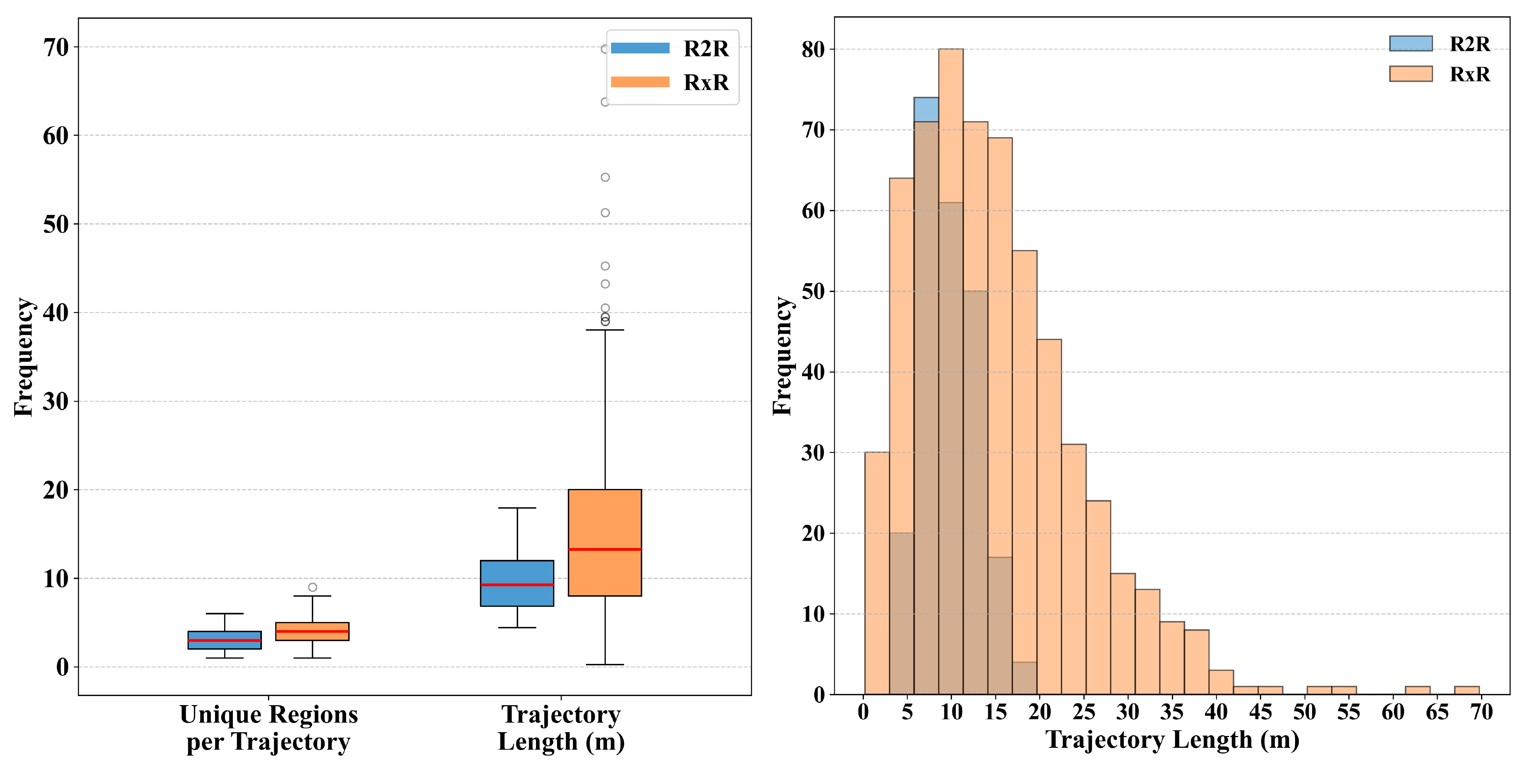}
    \vspace{-2em}
    \caption{Dataset comparison between FloorPlan-R2R Val-Seen and FloorPlan-RxR on Val-Seen split.}
    \vspace{-1em}
    \label{fig:r2r_rxr}
\end{figure}
\textbf{Results on FloorPlan-RxR.} Except for FP-Nav-v-rxr, all models are evaluated in a zero-shot manner on the FloorPlan-RxR dataset.
As shown in Table~\ref{tab:main_results}, the results exhibit trends consistent with those observed on FloorPlan-R2R, demonstrating strong cross-dataset generalization.
Since FloorPlan-RxR contains longer trajectories that traverse more regions, the task is inherently more challenging, leading to a performance drop across all methods.
Nevertheless, FP-Nav-v-rxr achieves further gains by adding long-horizon trajectories, highlighting both the model’s scalability and the effectiveness of our proposed learning strategy.

The results in Table~\ref{tab:main_results} show that both Qwen-zs and Navid-zs achieve higher performance on the FloorPlan-RxR dataset than on FloorPlan-R2R for both seen and unseen splits.
According to our statics, we find that FloorPlan-RxR contains a larger proportion of short trajectories, as illustrated in Figure~\ref{fig:r2r_rxr}.
The box plot indicates that FloorPlan-RxR has a lower minimum trajectory length, while the histogram reveals that it includes over 70 trajectories shorter than 5 meters, significantly more than FloorPlan-R2R.
These simpler trajectories increase the chance of navigation success.
For instance, on the Val-Seen split of FloorPlan-RxR, Navid-zs successfully completes 37 trajectories, among which 23 (about 62\%) are shorter than 5 meters.
In contrast, FloorPlan-R2R contains only 18 such short trajectories, explaining why zero-shot models perform better on FloorPlan-RxR dataset.

Conversely, the finetuned models exhibit lower performance on FloorPlan-RxR than on FloorPlan-R2R.
This is primarily due to the distribution shift between the two datasets, as shown in Figure~\ref{fig:r2r_rxr}.
Trajectories in FloorPlan-RxR are longer and more diverse, showing a right-skewed distribution with outliers reaching up to 70 meters, representing challenging, long-horizon navigation cases that are difficult to complete.
FloorPlan-R2R, in contrast, consists of shorter and more uniformly distributed trajectories, reflecting its relatively constrained navigation complexity.
Furthermore, FloorPlan-RxR trajectories typically traverse more regions, adding additional reasoning challenges.

Because Navid-ft, FP-Nav, and FP-Nav-v are fine-tuned only on FloorPlan-R2R, they struggle to generalize to the longer and more complex trajectories in FloorPlan-RxR.
By contrast, FP-Nav-v-rxr, which is trained with an additional 1.7k long-horizon episodes from FloorPlan-RxR, achieves clear improvements, confirming the benefits of augmenting with longer trajectories.
Nevertheless, since these additional episodes are still fewer than the 9k episodes used in FloorPlan-R2R, its performance remains slightly lower than on FloorPlan-R2R.
\begin{figure*}[t]
    \centering
    \includegraphics[width=1.0\linewidth, trim=0 1220 0 20, clip]{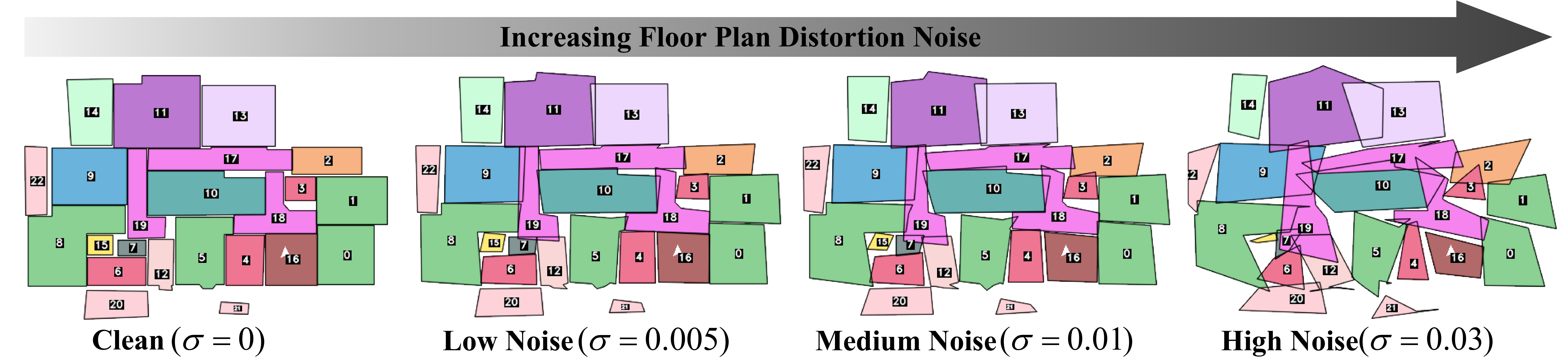}
    \vspace{-2em}
    \caption{Visualization of floor plan distortion under varying levels of geometric jitter.}
    \vspace{-1em}
    \label{fig:floorplan_distortion_noise}
\end{figure*}

\begin{figure}[t]
    \centering
    \includegraphics[width=1.0\linewidth, trim=0 800 0 0, clip]{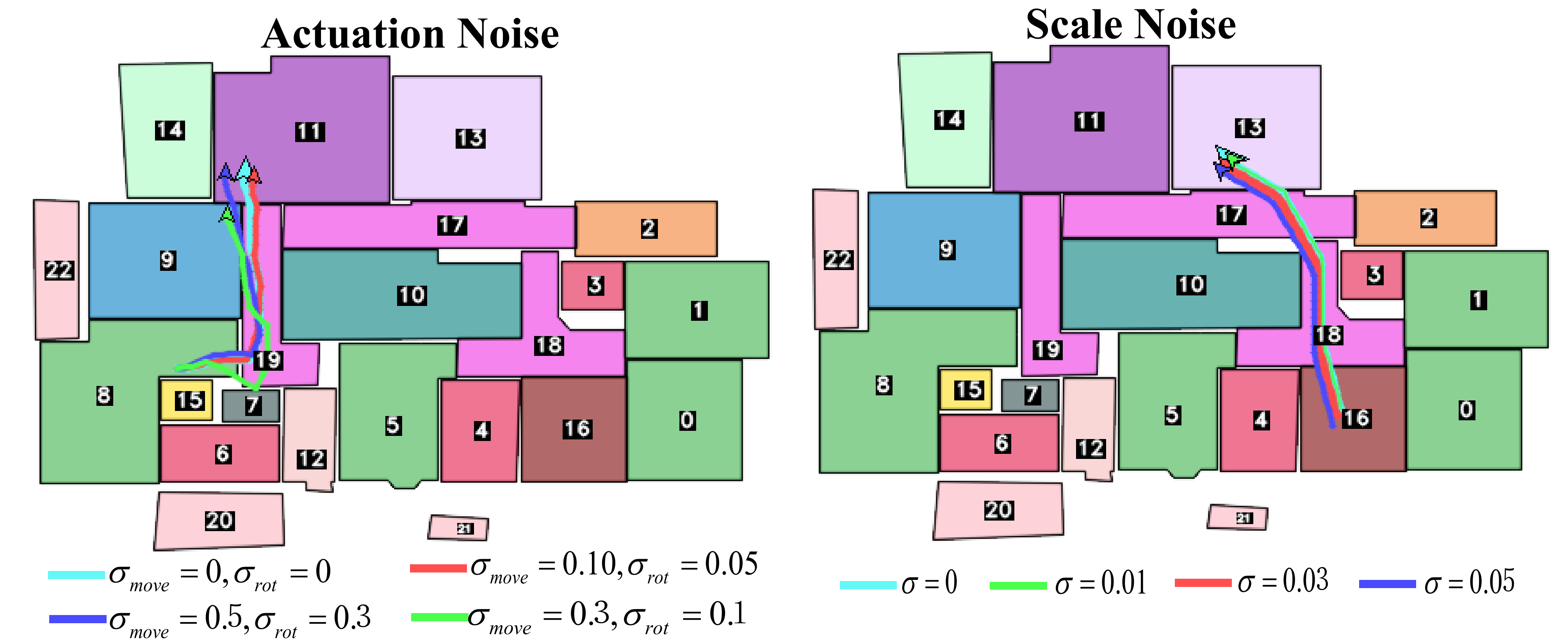}
    \vspace{-2em}
    \caption{Visualization of actuation noises and scale noises.}
    \label{fig:actuation_scale_noise}
\end{figure}

\begin{table}[h]
\centering
\vspace{-1.5em}
\caption{Robustness to actuation, floor plan scale, and floor plan distortion noises.}
\vspace{-1em}
\label{tab:noise}
\renewcommand{\arraystretch}{1.3}
\resizebox{\linewidth}{!}{
\begin{tabular}{C{0.5cm} C{3.5cm} C{0.9cm} C{0.9cm} C{0.9cm} C{0.9cm}}
\hline\hline
\rowcolor{gray!30}\multicolumn{6}{c}{\textbf{Experiment 1: Robustness to actuation noises}} \\
\hline
\# & Noise & NE$\downarrow$ & OSR$\uparrow$ & \textbf{SR}$\uparrow$ & \textbf{SPL}$\uparrow$\\
\hline
1 & $\sigma_{\text{move}}=0.00,\sigma_{\text{rot}}=0.00$ & 9.8 & 35.2 & 20.9 & \textbf{15.3} \\
2 & $\sigma_{\text{move}}=0.10,\sigma_{\text{rot}}=0.05$ & \textbf{8.4} & \textbf{42.4} & \textbf{22.0} & 14.5 \\
3 & $\sigma_{\text{move}}=0.30,\sigma_{\text{rot}}=0.10$ & 8.9 & 39.5 & 21.4 & 11.5 \\
4 & $\sigma_{\text{move}}=0.50,\sigma_{\text{rot}}=0.30$ & 8.60 & 37.1 & 17.5 & 6.7 \\

\hline\hline
\rowcolor{gray!30}\multicolumn{6}{c}{\textbf{Experiment 2: Robustness to floor plan scale noises}} \\
\hline
\# & Noise & NE$\downarrow$ & OSR$\uparrow$ & \textbf{SR}$\uparrow$ & \textbf{SPL}$\uparrow$\\
\hline
1 & $\sigma_{\text{scale}}=0.00$ & \textbf{9.8} & \textbf{35.2} & \textbf{20.9} & \textbf{15.3} \\
2 & $\sigma_{\text{scale}}=0.01$ & 10.2 & 32.4 & 18.6 & 12.8 \\
3 & $\sigma_{\text{scale}}=0.03$ & 10.0 & 32.6 & 19.6 & 13.7 \\
4 & $\sigma_{\text{scale}}=0.05$ & 10.1 & 32.4 & 18.3 & 13.0 \\

\hline\hline
\rowcolor{gray!30}\multicolumn{6}{c}{\textbf{Experiment 3: Robustness to floor plan distortion noises}} \\
\hline
\# & Noise & NE$\downarrow$ & OSR$\uparrow$ & \textbf{SR}$\uparrow$ & \textbf{SPL}$\uparrow$\\
\hline
1 & $\sigma_{\text{jitter}}=0.000$ & \textbf{9.8} & \textbf{35.2} & \textbf{20.9} & \textbf{15.3} \\
2 & $\sigma_{\text{jitter}}=0.005$ & 10.0 & 34.9 & 20.1 & 14.0 \\
3 & $\sigma_{\text{jitter}}=0.010$ & 9.8 & 33.5 & 20.1 & 14.3 \\
4 & $\sigma_{\text{jitter}}=0.030$ & 9.8 & 33.9 & 18.5 & 13.3 \\
\hline
\end{tabular}}
\vspace{-2em}
\end{table}

\subsection{Robustness to Actuation and Floor Plan Noises}
We systematically test FP-Nav's robustness by increasing the intensity of the noises defined in \S~\ref{sec:noise}.

\textbf{Robustness to Actuation Noises:} Figure~\ref{fig:actuation_scale_noise} illustrates the effect of different levels of actuation noise on the same action sequence. 
While the drift at each step is small, the accumulated drift can become significant over time. 
The results presented in Table~\ref{tab:noise} demonstrate that our model exhibits remarkable resilience to actuation noise. 
Despite the cumulative drift induced by $\sigma_{move}$ and $\sigma_{drift}$, the Success Rate (SR) remains relatively stable. 
Moreover, with minimal action perturbation, performance even shows a slight improvement. 
These findings confirm that our method does not overfit to the oracle poses and is relatively robust to small imperfections in actuation.

\textbf{Generalization to Imperfect Floor Plans:} As illustrated in Figure~\ref{fig:floorplan_distortion_noise}, increasing distortion noise results in a progressively irregular floor plan. 
Figure~\ref{fig:actuation_scale_noise} further demonstrates the deviation in the trajectory caused by noise in the floor plan scale. 
Despite the presence of noise, the results from Experiment 2 and 3 in Table~\ref{tab:noise} indicate that inaccuracies in scaling and geometric perturbations lead to only a modest decline in navigation performance. 
These findings suggest that the agent does not rely on absolute coordinate matching. 
Instead, it appears to have learned to leverage invariant spatial layouts. 
This robustness provides a compelling bridge between high-precision simulators and real-world deployment, where floor plans are often informal or hand-drawn.

\subsection{Ablation Study}
\begin{figure}[t]
    \centering
    \includegraphics[width=1.0\linewidth, trim=0 10 50 0, clip]{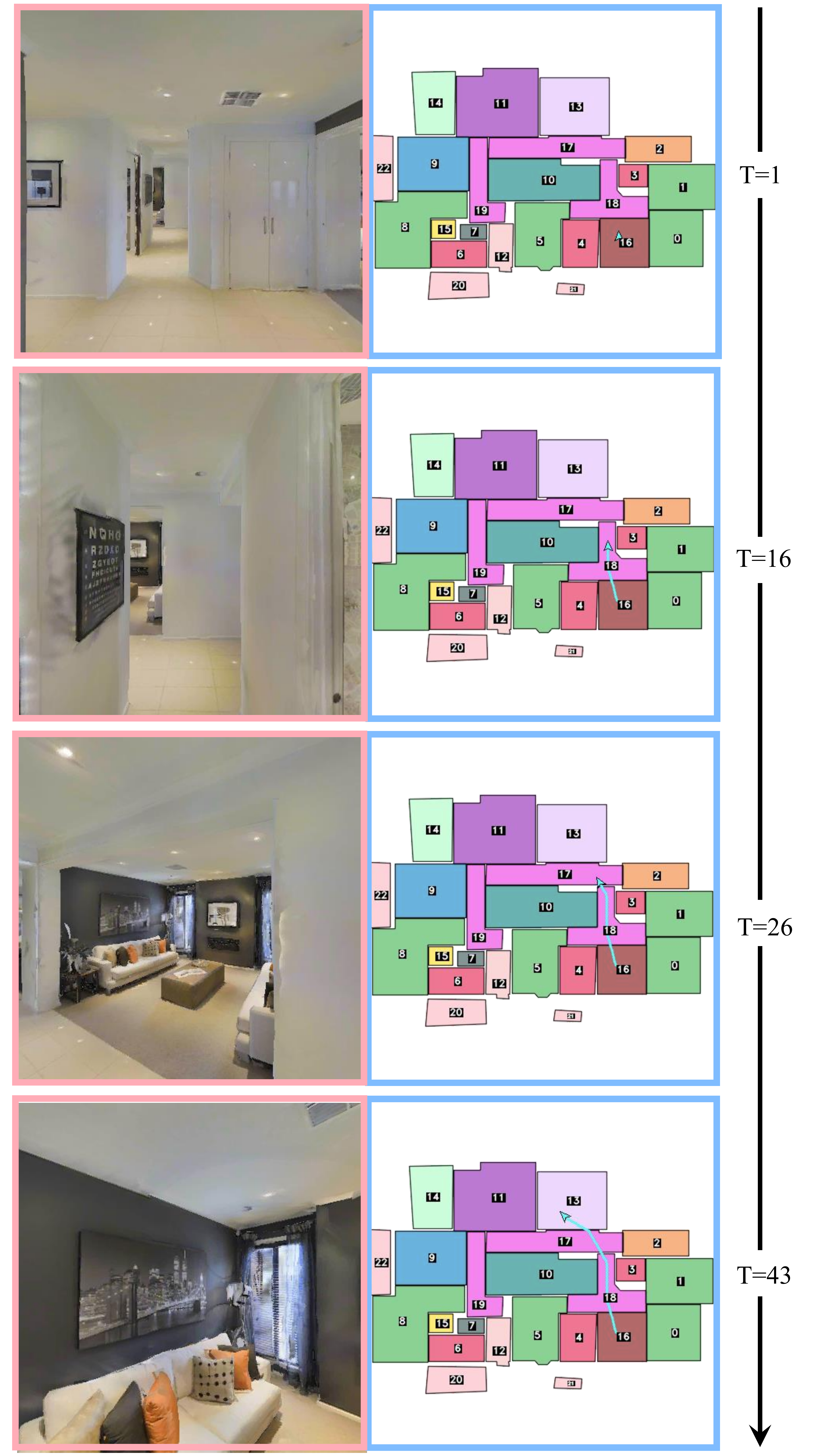}
    \vspace{-2em}
    \caption{An example of a successful episode using the spatio-temporally aligned input strategy. At each step, the agent’s pose is plotted on the floor plan and concatenated with the corresponding egocentric observation to form a synchronized dual-view input sequence.}
    \label{fig:visualization}
    \vspace{-2em}
\end{figure}

\textbf{Is the proposed spatio-temporally aligned video stream effective?}\label{ablation:input} 
Figure~\ref{fig:visualization} visualizes a successful navigation episode under the proposed spatio-temporally aligned input strategy. 
A natural question then arises: Is this dual-view input strategy truly more effective than other input designs introduced in Figure~\ref{fig:input_strategies}?
To answer this question, we evaluate the four input strategies and present the results in Table~\ref{tab:input_strategy}.

\begin{table}[h]
\vspace{-1em}
\centering
\caption{Ablation study on different input strategies.}
\vspace{-1em}
\label{tab:input_strategy}
\resizebox{\linewidth}{!}{
\begin{tabular}{C{0.5cm} C{3.5cm} C{0.9cm} C{0.9cm} C{0.9cm} C{0.9cm}}
\toprule
\# & Method & NE$\downarrow$ & OSR$\uparrow$  & \textbf{SR}$\uparrow$ & \textbf{SPL}$\uparrow$\\
\midrule
1 & static separate input & 9.8 & 3.9 & 2.5 & 2.3 \\
2 & dual-stream temporal fusion & 9.8 & 9.8 & 4.2 & 2.4 \\
3 & interleaved frame fusion & \textbf{9.6} & 4.4 & 2.6 & 2.4 \\
4 & spatio-temporally aligned & 9.8 & \textbf{35.2} & \textbf{20.9} & \textbf{15.3} \\
\bottomrule
\end{tabular}}
\end{table}

We first provide the model with a static floor plan image and an observation video as two separate inputs. 
In this case, the agent must infer correspondences between dynamic egocentric views and a static layout, yielding weak performance due to the lack of temporal synchronization.

Next, we construct a floor plan navigation video that visualizes the agent’s pose and trajectory over time, and feed the two streams into the model separately (Figure~\ref{fig:input_strategies} b). 
Although this method provides explicit localization cues and slightly improves OSR and SR, the separate processing prevents precise frame-level alignment.

We then test an interleaved strategy where observation frames and floor plan frames at the same time step are alternately fed into the model. 
Although this design aims to couple temporally corresponding views, it breaks the temporal continuity of each stream, leading to weaker integration of observation and floor plans and degraded performance.

In contrast, our spatio-temporally synchronized dual-view strategy performs simple yet effective early fusion: each observation frame is concatenated with its corresponding floor plan view of the agent’s pose, forming a single aligned image.
This preserves spatial–temporal continuity, enabling the model to perceive local details and global structure jointly, thus achieving better alignment and the best performance.

\textbf{Does the model learn to align the floor plan with egocentric observations?} We conduct a series of experiments to verify whether the model truly relies on the floor plan during navigation, rather than focusing solely on egocentric observations. 
As shown in Table~\ref{tab:floorplan-usage}, masking increasing portions of the floor plan leads to consistent drops in SR and SPL and a rise in NE, indicating reduced spatial reference forces the agent to explore more.
In some cases, the agent still passes the correct goal but fails to stop, explaining the slight OSR increase.
When the floor plan is fully masked, SR drops by 41\%, confirming that the model indeed depends on floor plan cues for navigation.
To further validate this, we provide mismatched floor plans from other scenes (Row 6 in Table~\ref{tab:floorplan-usage}).
Compared to the correct floor plan, SR and SPL drop by 56\% and 64\%, respectively, showing that inaccurate spatial priors severely mislead navigation.
This confirms that the model leverages the floor plan for global reasoning, rather than relying solely on observations.

\begin{table}[t]
\centering
\vspace{-6pt}
\caption{Investigation of whether the model truly learns to use floor plans during navigation.}
\label{tab:floorplan-usage}
\vspace{-1em}
\resizebox{\linewidth}{!}{
\begin{tabular}{C{0.5cm} C{3.cm} C{0.9cm} C{0.9cm} C{0.9cm} C{0.9cm}}
\toprule
\# & Method & NE$\downarrow$ & OSR$\uparrow$  & \textbf{SR}$\uparrow$ & \textbf{SPL}$\uparrow$\\
\midrule
1 & full floorplan & \textbf{9.8} & \textbf{35.2} & \textbf{20.9} & \textbf{15.3}\\
2 & mask 25\% & 10.1 & 28.7 & 17.7 & 13.4 \\
3 & mask 50\% & 10.0 & 29.8 & 16.8 & 13.0 \\
4 & mask 75\% & 10.2 & 30.7 & 14.4 & 10.5 \\
5 & mask 100\% & 10.8 & 32.0 & 12.3 & 8.0 \\
6 & random floorplan & 10.7 & 34.4 & 9.1 & 5.5 \\ 
\bottomrule
\end{tabular}}
\vspace{-2em}
\end{table}

\textbf{What is the impact of different auxiliary tasks.} 
Since action-based QA supervision alone is insufficient to elicit floor plan understanding in MLLMs for VLN, we introduce three auxiliary tasks to enhance the alignment between floor plans, egocentric observations and instructions.

\begin{table}[h]
\centering
\vspace{-6pt}
\caption{Effect of auxiliary tasks on navigation performance.}
\vspace{-1em}
\label{tab:auxiliary-tasks}
\resizebox{\linewidth}{!}{
\begin{tabular}{C{0.5cm} C{3.cm} C{0.9cm} C{0.9cm} C{0.9cm} C{0.9cm}}
\toprule
\# & Method & NE$\downarrow$ & OSR$\uparrow$  & \textbf{SR}$\uparrow$ & \textbf{SPL}$\uparrow$\\
\midrule
1 & w/o aux tasks & 10.0 & 30.5 & 17.8 & 13.0 \\
2 & RL & 9.8 & 32.0 & 19.1 & 12.8 \\
3 & RL + TR & \textbf{9.7} & 32.0 & 20.1 & 14.5 \\
4 & RL + TR + IS & 9.8 & \textbf{35.2} & \textbf{20.9} & \textbf{15.3} \\
\bottomrule
\end{tabular}}
{\footnotesize
\textit{Abbr: RL = Region Localization, TR = Trajectory Reasoning, IS = Instruction Summarization.}
}
\vspace{-1em}
\end{table}

Table~\ref{tab:auxiliary-tasks} summarizes the effects of each auxiliary task.
Removing all of them causes about a 15\% drop in performance, confirming their effectiveness.
Region localization improves SR and OSR by helping the agent infer the region type of its current location.
Trajectory reasoning further enhances all metrics by promoting region-level planning.
Finally, instruction summarization aligns concise language with navigation videos, achieving the best overall results.
Together, these tasks progressively strengthen spatial alignment, instruction understanding, and navigation planning.

\textbf{The effect of training with different parameters.} The substantial improvement achieved by unfreezing the vision encoder naturally raises a question: what is the impact of unfreezing different components of Qwen?

\begin{table}[h]
\centering
\small
\vspace{-6pt}
\caption{Effect of training different model components on navigation performance.}
\label{tab:parameters}
\vspace{-1em}
\resizebox{\linewidth}{!}{
\begin{tabular}{C{0.5cm} C{3.cm} C{0.9cm} C{0.9cm} C{0.9cm} C{0.9cm}}
\toprule
\# & Method & NE$\downarrow$ & OSR$\uparrow$  & \textbf{SR}$\uparrow$ & \textbf{SPL}$\uparrow$\\
\midrule
1 & MLP & 10.4 & 3.2 & 2.3 & 1.9 \\
2 & MLP + LLM & 9.8 & 35.2 & 20.9 & 15.3 \\
3 & MLP + LLM + Vision & \textbf{9.6} & \textbf{37.4} & \textbf{25.9} & \textbf{21.7} \\
\bottomrule
\end{tabular}}
\end{table}

Table~\ref{tab:parameters} presents the results of this analysis. 
When only the MLP projection head is fine-tuned, performance even drops below the Qwen-zs baseline in Table~\ref{tab:main_results}. 
This outcome is expected, since the MLP serves merely as a shallow adapter that maps visual embeddings into the embedding space of the LLM, and such a two-layer network lacks the capacity to learn new skills such as understanding the floor plan or reasoning about navigation.
Allowing both the MLP and LLM to update leads to a clear performance gain, suggesting that the LLM can already learn to interpret concise navigation instructions when conditioned on floor plan. 
Finally, unfreezing all parameters achieves the best results, as the full capacity of the MLLM is utilized to jointly refine perception, alignment, and reasoning.

\begin{figure*}[t]
    \centering
    \includegraphics[width=0.95\linewidth, trim=0 600 0 0, clip]{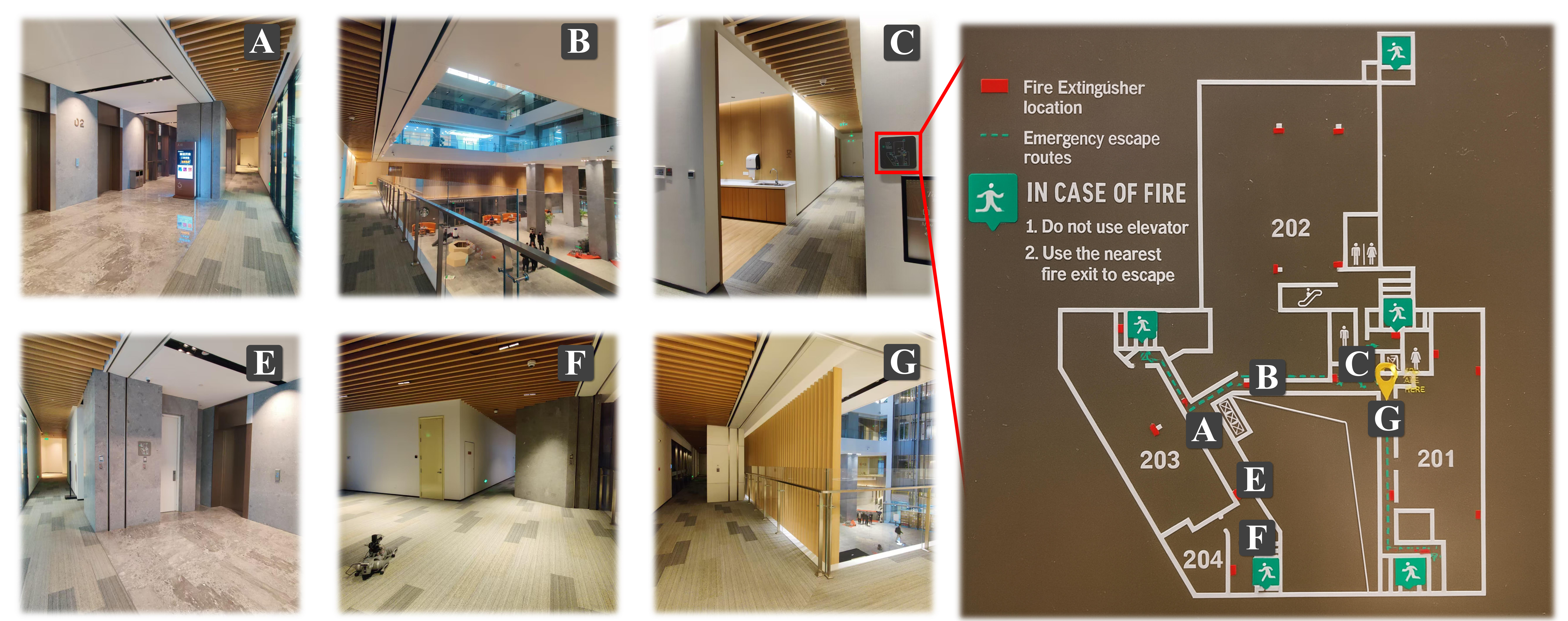}
    \vspace{-1em}
    \caption{Real-world experimental scenes and the corresponding floor plan.}
    \vspace{-1em}
    \label{fig:real_scenes}
\end{figure*}
\section{Real-World Experiments}
\label{sec:real-world experiments}
We conduct real-world experiments to evaluate whether our proposed FP-Nav can follow concise instructions and utilize floor plans to complete long-horizon navigation tasks that need to cross multiple regions.
Prior real-world VLN evaluations typically take place in single room environments such as meeting rooms, offices or laboratories~\cite{zhang2024navid, cheng2024navila}.
In these settings, trajectories are short, region transitions are minimal, and fine-grained instructions compensate for the lack of global spatial priors.
In contrast, our setting reflects a substantially more challenging and realistic scenario.
As shown in Figure~\ref{fig:real_scenes}, we conduct real-world experiments in a large building which spans approximately 1,370 m², containing 25 regions including offices, corridors, meeting rooms, and bathrooms. 
Navigating in such a layout is challenging which requires the agent to traverse multiple regions, reason over wide-range spaces, and follow concise instructions.

\subsection{Real-world Deployment Details}
\label{sec:real-world-implementation-details}
\begin{figure*}[t]
    \centering
    \includegraphics[width=1.0\linewidth, trim=0 1050 0 0, clip]{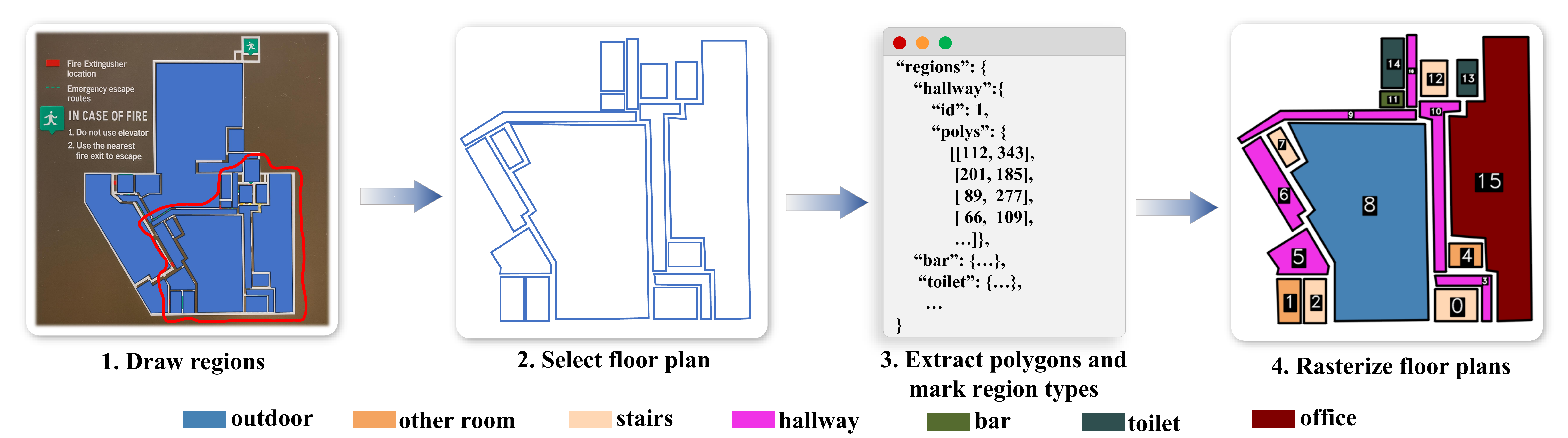}
    \vspace{-2em}
    \caption{The pipeline of generating real-world floor plan.}
    \label{fig:real_floorplan}
    \vspace{-1em}
\end{figure*}

To generate floor plans compatible with the standardized representation used during training, we design a semi-automatic conversion pipeline, as illustrated in Figure~\ref{fig:real_floorplan}. 
Starting with a raw photo of the building's floor plan (Figure~\ref{fig:real_scenes}), we manually trace region contours in PowerPoint to delineate region boundaries. 
We then select the appropriate floor plan based on the experimental environment. 
The pipeline automatically extracts the polygon vertices of each region and stores them in a structured JSON file containing region polygons and unique identifiers. 
Region types are manually labeled at this stage. 
Finally, the pipeline automatically rasterizes the floor plan images in the FloorPlan-VLN format, ensuring domain consistency between simulation and real-world deployment. 
This process is straightforward, requires minimal manual effort, and generates floor plans directly compatible with FP-Nav’s input. 
Additionally, hand-drawn maps can also be converted using this pipeline, as demonstrated in Figure~\ref{fig:hand2floor}.

Next, we develop a human-robot interaction interface that enables users to mark any starting position and heading on the floor plan as the robot’s initial pose. 
The user then provides a concise instruction following the FloorPlan-VLN format (start region, target region, stopping condition) and the robot subsequently executes navigation based on the concise instruction and specified pose.

While simulation provides perfect pose information for trajectory rendering, the real robot relies on LiDAR-based odometry to estimate pose changes over time. 
We integrate these odometry updates, apply coordinate transforms anchored at the user-specified start pose, and project the resulting trajectory onto the floor plan using a precomputed scale factor derived from the building’s layout. 
This enables real-time visualization of the robot’s movement on the floor plan and mirrors the pipeline used during training.

\begin{figure}[t]
    \centering
    \includegraphics[width=1.\linewidth, trim=100 130 100 0, clip]{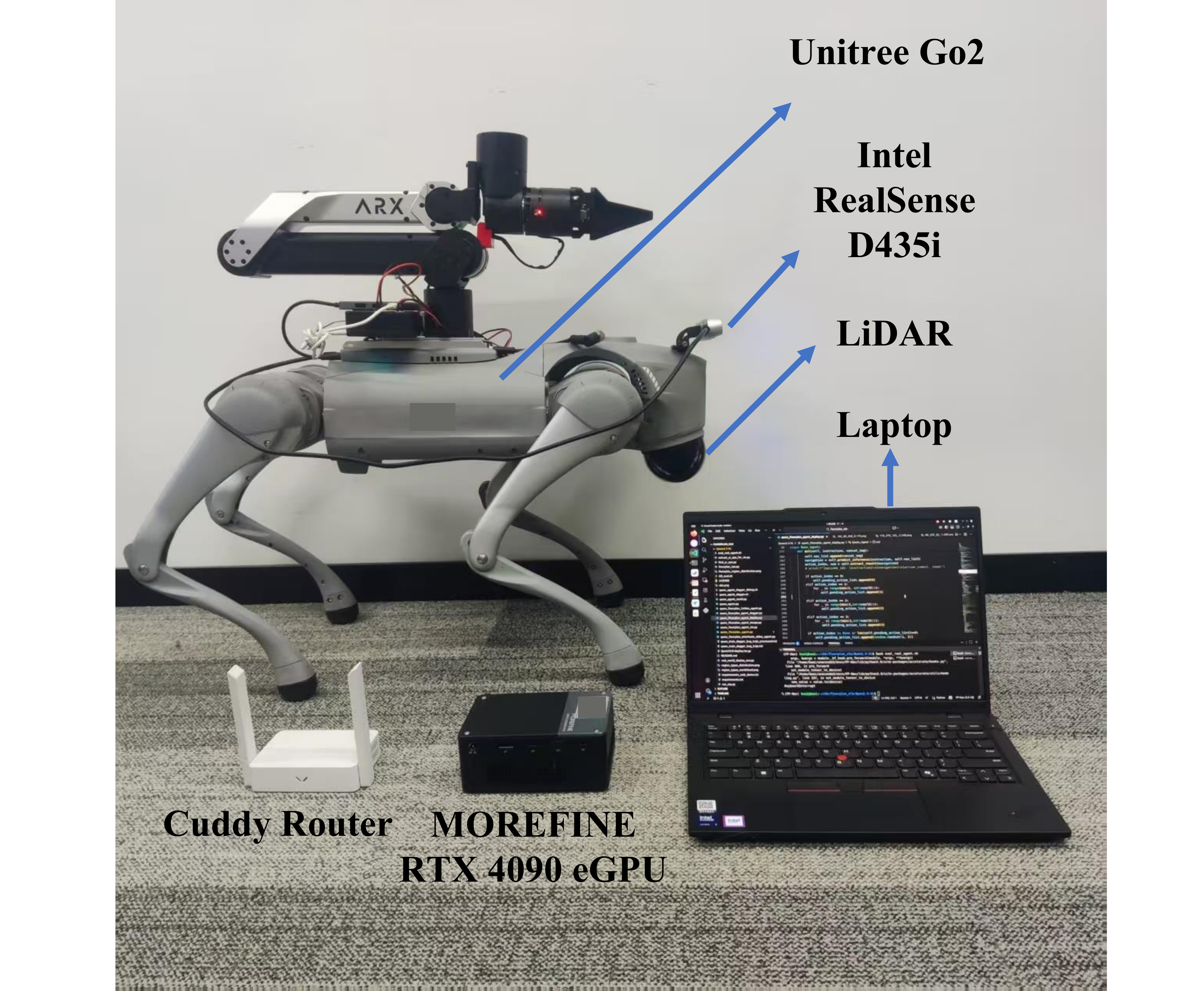}
    \begin{minipage}{1.\linewidth}
    \caption{Real-world robot setup.}
    \label{fig:real_robot}
    \end{minipage}
    \vspace{-2em}
\end{figure}

As shown in Figure~\ref{fig:real_robot}, we deploy FP-Nav on a Unitree Go2 quadruped robot equipped with a Jetson Orin, an Intel RealSense D435i camera, and a LiDAR for odometry. 
Robot perceptions derived from RealSense are send to an external laptop with an Morefine RTX 4090 eGPU (16GB) to perform FP-Nav inference, and actions are transmitted back to the robot via ZeroMQ. 
Locomotion execution uses the high-level control APIs provided by Unitree, enabling stable motion while focusing evaluation solely on navigation decision-making rather than low-level control.

\subsection{Results}
\vspace{-4pt}
We evaluate Navid-ft and FP-Nav-v-rxr at six locations (A–F) shown in Figure~\ref{fig:real_scenes}, testing 5 trajectories among these locations, with each test repeated 5 times, with slightly different start poses and stop conditions, resulting in a total of 25 episodes.
Each episode crosses at least 2 regions and the average length is about 12.2 meters which is a challenging task setting.
An episode is considered successful if the robot stops within 3 meters of the target region and we use standard metrics SR and NE, the results are shown in Table~\ref{tab:real-exp}.

As navigating long-horizon trajectories in the real world compounds early actuation errors and visual domain shifts, zero-shot deployment remains highly challenging. 
Consequently, the baseline Navid-ft struggles significantly, achieving a success rate of only 8.0\% (2 successful episodes). 
In stark contrast, our FP-Nav-v-rxr achieves a success rate of 24.0\% (6 successful episodes), demonstrating a notable relative improvement.
More importantly, despite the absence of any real-world fine-tuning and the presence of physical actuation noise, the 24.0\% real-world SR of FP-Nav closely matches its simulation performance on the val-unseen split (28.8\%, as shown in Table~\ref{tab:main_results}). 
This indicates that our method is robust against actuation variances and floor plan perturbations. 
These results validate the feasibility and potential of FloorPlan-VLN.

\begin{table}[t]
\centering
\small
\caption{The results of real-world experiments.}
\label{tab:real-exp}
\vspace{-0.7em}
\resizebox{\linewidth}{!}{
\begin{tabular}{C{1cm}C{3.0cm}C{1.cm}C{1.cm}}
\toprule
\# & Method & NE$\downarrow$ & SR$\uparrow$\\
\midrule
1 & Navid-ft & 9.3 & 8.0 \\
2 & FP-Nav-v-rxr & 6.4 & 24.0 \\
\bottomrule
\end{tabular}}
\vspace{-1em}
\end{table}

\section{Conclusion}
\vspace{-2pt}
\label{sec:conclusion}
We introduced FloorPlan-VLN, a novel paradigm that shifts Vision-Language Navigation toward human-like spatial reasoning by utilizing floor plans as global priors under concise instructions. 
To support this, we contributed a standardized dataset and proposed FP-Nav, a strong MLLM-based baseline featuring spatio-temporally aligned inputs and auxiliary reasoning tasks. 
Notably, our framework demonstrates strong zero-shot robustness against simulated actuation noises and real-world physical variances, proving its practical deployment potential.
Beyond current achievements, a critical insight from our work is that spatial diagrams act as a universal cognitive bridge between natural language and physical environments.
Future efforts could focus on enabling agents to actively refine their internal cognitive maps when encountering key spatial structures in the real world, while also enhancing the ability of MLLMs to reason with floor plans.

\bibliographystyle{IEEEtran}
\bibliography{reference}

\vfill

\end{document}